\documentclass[sort&compress,journal]{IEEEtran} 
\usepackage{amsmath,amsfonts,amssymb}
\usepackage{algorithmic}
\usepackage{algorithm}
\usepackage{array}
\usepackage[caption=false,font=normalsize,labelfont=sf,textfont=sf]{subfig}
\usepackage{textcomp}
\usepackage{stfloats}
\usepackage{url}
\usepackage{verbatim}
\usepackage{graphicx}
\usepackage{cite}
\usepackage{hyperref}
\hypersetup{hidelinks,colorlinks=true,allcolors=black,pdfstartview=Fit,breaklinks=true}

\begin{document}

\title{Distributed Control for a Robotic Swarm to Pass through a Curve Virtual Tube}

\author{Quan Quan, Yan Gao, Chenggang Bai
\thanks{Quan Quan, Yan Gao and Chenggang Bai and are with the School of Automation Science and Electrical Engineering, Beihang University, Beijing 100191, P. R. China (email:  qq\_buaa@buaa.edu.cn; buaa\_gaoyan@buaa.edu.cn; bcg@buaa.edu.cn)}}




\maketitle

\begin{abstract}
Robotic swarm systems are now becoming increasingly attractive for many challenging applications. The main task for any robot is to reach the destination while keeping a safe separation from other robots and obstacles. In many scenarios, robots need to move within a narrow corridor, through a window or a doorframe. In order to guide all robots to move in a cluttered environment, a curve virtual tube with no obstacle inside is carefully designed in this paper. There is no obstacle inside the tube, namely the area inside the tube can be seen as a safety zone. Then, a distributed swarm controller is proposed with three elaborate control terms: a line approaching term, a robot avoidance term and a tube keeping term. Formal analysis and proofs are made to show that the curve virtual tube passing through problem can be solved in a finite time. For the convenience in practical use, a modified controller with an approximate control performance is put forward. Finally, the effectiveness of the proposed method is validated by numerical simulations and real experiments. To show the advantages of the proposed method, the comparison between our method and the control barrier function method is also presented in terms of calculation speed.  
\end{abstract}

\begin{IEEEkeywords}
Distributed control, robotic swarm, vector field, artificial potential field, virtual tube.
\end{IEEEkeywords}

\section{Introduction}
\IEEEPARstart{R}{ecently}, advances in research have brought robotic swarm systems to a level of sophistication which makes it increasingly attractive for a variety of complex applications like sensing, mapping, search and rescue \cite{Chung(2018)}. In order to make these applications a reality, it is necessary for the robotic swarm to have the ability to operate in a cluttered environment and reach the appointed destination in a distributed way. Moving within a narrow corridor, through a window or a doorframe is a very common scenario \cite{ding2021epsilon}. In this process, not only should each robot avoid collisions with obstacles, but all robots also need to avoid collisions with each other.

\begin{figure}
	\centering
	\includegraphics[width=\columnwidth]{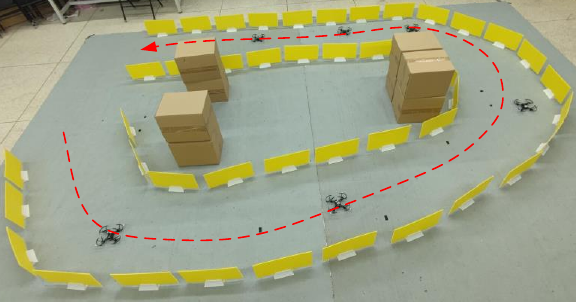}
	\caption{A curve virtual tube is designed for guiding a multicopter swarm in a cluttered environment.}
	\label{First}
\end{figure}


Many successful methods have been put forward for planning and controlling multiple robots in a cluttered environment. The typical methods can mainly be classified into three types: formation, multi-agent trajectory planning and control-based methods. The \emph{formation} here refers specifically to all robots keeping a geometry structure as a whole, which suits both semi-autonomous and fully-autonomous control \cite{ren2008distributed,Cheung(2019),Oh(2015),Khan(2016),Zhao(2019),Saska(2020),dong2020controlling}. Each robot in the formation usually remains a prespecified pose and makes the formation stable and robust. The pilot or the control program can control the formation to avoid the obstacles in the environment, which is like controlling a single robot. The inter-robot collision avoidance is mainly achieved by the fixed geometry. When the formation operates in a cluttered environment and needs to pass through some narrow spaces or corridors, necessary transformations have to be carried out \cite{Zhao(2019)}. The affine formation maneuver control is especially suitable for the transformation control \cite{xu2020affine}. However, the formation is not perfect in all circumstances and the main weakness is its limited scalability and adaptability. If there are hundreds or even thousands of robots, the increase in quantity leads to the expansion of the physical size and the formation will be too big to maintain feasibility \cite{Hao(2016)}. Besides, when some robots need to change their locations, it may cause chaos in the formation and the controller will become very complex. 


The \emph{multi-agent trajectory planning} produces collision-free trajectories with higher-order continuity for all robots in a centralized or distributed way \cite{Mellinger(2012),Augugliaro(2012),luo2019importance,morgan2016swarm,Luis(2019),Park(2020),Zhou(2020)}. Different from the velocity or acceleration command of the formation control, the higher-order trajectory, such as Bezier curves \cite{Park(2020)} and B-splines \cite{Zhou(2020)}, can get a better control performance especially for the multicopter. Given a specific target point, any robot should find a discrete geometric path in the global map first and then optimize the path to a feasible trajectory locally with no conflict with obstacles and other robots' trajectories \cite{Ding(2019)}. However, when there are multiple robots operating densely, trajectories planning becomes inappropriate intuitively and infeasible in real practice, the reason of which is the sharp increase in computational complexity and decrease of the feasible region \cite{Augugliaro(2012)}, \cite{chen2016pomdp}. Besides, distributed multi-agent trajectory planning needs any robot to share its planned trajectory with others via wireless communication, which brings a huge communication pressure when the number of robots increases \cite{Zhou(2020)}. The communication uncertainties, such as broadcast delay and packet loss, can also make the trajectory optimization infeasible under some circumstances \cite{Tahir(2019)}. 


For robotic \emph{swarm} navigation and control, the control-based methods are widely used because of their simplicity and accessibility \cite{wolf2008artificial}. Control-based methods usually use a simple controller to react to obstacles or other robots, which have a good quality to achieve a fast and reactive response to a dynamic environment and a low demand for computation and communication resources. Different from the multi-agent trajectory planning, the \emph{control-based methods} directly guide the robots' movement with the velocity or acceleration command according to the global path and current local information \cite{Miao(2016),Wang(2017),Quan(2021)}. Although the control-based methods possess a weaker control performance compared with the multi-robot trajectory planning  \cite{Luis(2019),Park(2020),Zhou(2020)}, they are more suitable for the large-scale robotic swarm, such as the potential-based method \cite{Quan(2021)} and the vector field method \cite{Miao(2016)}, which directly guide the robots' movement according to the global path and current local information \cite{Wang(2017)}.  Besides, the control barrier function (CBF) method is also popular in recent years, which is summarized as a quadratic programming (QP) problem with better performance and higher demand on the computational resources \cite{Wang(2017)}. 

The method proposed in this paper is a modified type of the artificial potential field (APF) method belonging to the control-based method. The APF method was first introduced for manipulator control by Khatib \cite{khatib1986real}, which can be also seen as a gradient vector field method. In contrast, there are also non-potential vector field methods with curls non-zero \cite{panagou2014motion}, \cite{panagou2016distributed}. Compared with the CBF method, the APF method is especially suitable for dealing with multi-objective compositions at the same time. The task of the multi-objective composition is rather complicated for the CBF method, as multiple hard safety constraints may cause no feasible solution. Existing literature is limited to the combination of similar and complementary objectives, such as collision avoidance and connectivity maintenance \cite{wang2016multi}. For the APF method, each control objective can be described as a potential function, either attractive or repulsive. By summing up all potential functions, the corresponding vector field is directly generated with a gradient operation. Nevertheless, inappropriate definitions of the potential field will cause various problems, in which the most serious is local minima \cite{hernandez2011convergence}. The local minima problem is the appearance of unexpected equilibrium points where the composite potential field vanishes. This problem has limited the extensive application of the APF method. In the literature, there exist many different approaches to deal with the local minima problem while ensuring convergence and safety. Some approaches are willing to keep the robots away from or make robots leave from these deadlock points. In \cite{rostami2019obstacle}, the local minima are avoided by orthogonal decomposition of the repulsive component in the direction of parallel and perpendicular to the attractive component. In \cite{antich2005extending}, \cite{ge2005queues}, small disturbances are imposed to make robots escape local minima. Another kind of solution is to improve the APF method with optimization approaches, such as the evolutionary artificial potential field method \cite{vadakkepat2000evolutionary}, in which the APF method is combined with genetic algorithms to derive optimal potential field functions. Besides, in \cite{kim1992real}, the harmonic potential functions are introduced, whose undesired equilibrium points are all saddle points. The Morse function has a similar property, that is, all undesired local minima disappears as $\delta$ increases \cite{rimon1990exact}, \cite{panagou2015distributed}.

Motivated by the current studies, a \textit{curve virtual tube} is presented for guiding the robotic swarm in a cluttered environment. The term ``virtual tube" appears in the AIRBUS’s Skyways project, which enables unmanned airplanes to fly over the cities \cite{AIRBUS}. In our previous work \cite{Quan(2021)}, the \textit{straight-line virtual tube} is proposed for the air traffic control as flight routes are usually composed of several line segments. There is no obstacle inside the virtual tube and the area inside can be taken as a safety zone. In other words, the robots in the virtual tube have no need to consider conflicts with obstacles and only need to guarantee no collision with each other and the tube boundary. In this paper, the concept of the virtual tube is generalized. Compared with the straight-line virtual tube, the proposed curve virtual tube is generated along a continuous generating curve and the radius or width of the virtual tube is mutable. Hence, the curve virtual tube is more suitable for guiding the robotic swarm to move within a narrow corridor, through a window or a doorframe. Some similar concepts have been proposed in the literature. As shown in Fig. \ref{First}, the concept of the curve virtual tube is similar to the lane for autonomous road vehicles in \cite{Rasekhipour(2016)}, \cite{luo2018porca} and the corridor for a multi-UAV system in \cite{Tony(2020)}. In \cite{Rasekhipour(2016)}, autonomous road vehicles are restricted to moving inside the lane with a potential-field-based model predictive path planning controller.  In \cite{Liu(2017)}, a trajectory generation method using a safe flight corridor is proposed for a single multicopter, which defines a piecewise Bezier curve within the corridor by utilizing the convex hull property. In \cite{Tony(2020)}, the authors proposed a multi-drone skyway framework called CORRIDRONE, which is one such architecture that generates corridors for point-to-point traversal of several drones.

For controlling the robotic swarm with our proposed curve virtual tube, two problems are summarized, namely \textit{curve virtual tube planning problem} and \textit{curve virtual tube passing through problem}. This paper only aims to solve the latter one. For the curve virtual tube planning problem, the generating curve can be automatically obtained by interpolating with several waypoints, which can be generated from the traditional sampling-based \cite{Likhachev(2004),Dolgov(2010),Harabor(2011),gammell2014informed,gammell2015batch}, or search-based methods \cite{Kavraki(1996)}, \cite{LaValle(1999)}. Another feasible approach is to utilize an existing trajectory as the generating curve, which performs like a ``teach-and-repeat" system \cite{Gao(2019)}. Then the curve virtual tube is generated by expanding the generating curve and there must be no conflict with obstacles. The curve virtual tube passing through problem is solved in this paper with a distributed gradient vector field method. It should be noted that there is no uncertainty considered in this paper, namely all robots are able to obtain the information clearly and execute the velocity command exactly. For the real practice with various uncertainties, the \emph{separation theorem} can be introduced, which is proposed in our previous work \cite{quan2021far}, to solve the uncertainty problem. In short, all uncertainties are considered in the design of the safety radius of the robot, which is beyond the scope of this paper. Here, two models for robots and the curve virtual tube are first proposed. The robot model is a single integral model and the curve virtual tube model is set up with the concept of the generating curve and the cross section. To ensure all robots can reach the finishing line, a novel \emph{line-integral Lyapunov function} is proposed. Then, the single panel method and a Lyapunov-like barrier function are proposed for restricting robots to moving inside the curve virtual tube and avoiding collision with each other. Based on these functions and methods, a distributed swarm controller with a saturation constraint is designed. For practical use, a modified swarm controller with a similar control effect is also presented. It is proved that all robots are able to pass through the curve virtual tube and there is no local minima with \emph{invariant set theorem} \cite[p. 69]{Slotine(1991)}.

The major contributions of this paper are summarized as follows.
\begin{enumerate}
	\item[(i)] Based on the straight-line virtual tube introduced in our previous work \cite{Quan(2021)}, the curve virtual tube is proposed along with some concepts strictly defined with proper mathematical forms. The curve virtual tube is especially suitable for guiding the robotic swarm (tens of thousands of robots) to move within a narrow corridor, through a window or doorframe. This work makes a significant advance over existing planning and formation methods. Also, this work opens a new way of planning from ``one-robot, one-dimensional curve'' to ``swarm-robot, two (three)-dimensional tube''.
	\item[(ii)] A distributed swarm controller is proposed for the curve virtual tube passing through problem. Compared with the straight-line virtual tube passing through problem introduced in \cite{Quan(2021)}, the curve virtual tube passing through problem in this paper is more complicated and more general. With some active detection devices, such as cameras or radars, the proposed controller can work autonomously without wireless communication and other robots' IDs. The proposed controller is also very simple and can be computed at a high frequency, which is more suitable for the commonly used single-chip microcomputer. The calculation time of finding feasible solutions for our method and the CBF method is compared in a simulation.
	\item[(iii)]  A formal proof is proposed to show that there is no collision among robots, and all robots can keep within the tube and pass through the finishing line without getting stuck. The key to the proof is the use of the single panel method, which ensures that the angle between the orientation of the virtual tube keeping term and the orientation of the line approaching term is always smaller than $90^{\circ}$.
\end{enumerate} 

The rest of this paper is organized as follows. Section II
presents two models for the robot and the curve virtual tube. Then the curve virtual tube passing through problem is formulated. Section III presents some preliminaries. Section IV provides a distributed controller for the problem. Simulations and experiments are provided for supporting the theoretical results in Section V. Finally, Section VI discusses future work and concludes the paper. 

\section{Problem Formulation}

\subsection{Robot Model}

The robotic swarm consists of $M$ mobile robots in $\mathbb{R}^2$. For the vector field design and analysis, each robot has a single integral holonomic kinematics
\begin{equation}
	\dot{\mathbf{{p}}}_{i} =\mathbf{v}_{\text{c},i}, \label{SingleIntegral}
\end{equation}
in which $\mathbf{v}_{\text{c},i}\in {{\mathbb{R}}^{2}}$, $\mathbf{p}_{i}\in {{\mathbb{R}}^{2}}$ are the velocity command and position of the $i$th robot, respectively. It should be noted that $\mathbf{v}_{\text{c},i}$ is just the vector field to be designed. In the following, $\mathbf{v}_{\text{c},i}$ is only called as the \emph{velocity command}. Besides, $v_{\text{m},i}>0$ is set as the maximum permitted speed of the $i$th robot. Hence it is necessary to make $\mathbf{v}_{\text{c},i}$ subject to a saturation function
\begin{align*}
	\mathbf{v}_{\text{c},i}&=\text{sa}{\text{t}}\left(\mathbf{v}^{\prime}_{\text{c},i},{v_{\text{m},i}}\right) \\
	&={{\kappa}_{{v_{\text{m},i}}}}\left(\mathbf{v}^{\prime}_{\text{c},i}\right)\mathbf{v}^{\prime}_{\text{c},i},
\end{align*}
where $\mathbf{v}^{\prime}_{\text{c},i}\in {{\mathbb{R}}^{2}}$ is the original velocity command and
\begin{equation*}
	\text{sa}{\text{t}}\left( \mathbf{v}^{\prime}_{\text{c},i},{v_{\text{m},i}}\right) \triangleq
	\begin{cases}
		\mathbf{v}^{\prime}_{\text{c},i}  & 	\left \Vert \mathbf{v}^{\prime}_{\text{c},i}\right \Vert \leq {v_{\text{m},i}}\\ 
		{v_{\text{m},i}}\frac{\mathbf{v}^{\prime}_{\text{c},i}}{\left \Vert \mathbf{v}^{\prime}_{\text{c},i}\right \Vert } & \left \Vert \mathbf{v}^{\prime}_{\text{c},i}\right \Vert  >{v_{\text{m},i}}
	\end{cases}
\end{equation*}
\begin{equation*}
	{{\kappa }_{{v_{\text{m},i}}}}\left( \mathbf{v}^{\prime}_{\text{c},i}\right) \triangleq
	\begin{cases}
		1 & \left \Vert \mathbf{v}^{\prime}_{\text{c},i}\right \Vert \leq {v_{\text{m},i}}\\ 
		\frac{{v_{\text{m},i}}}{\left \Vert \mathbf{v}^{\prime}_{\text{c},i}\right \Vert } & 	\left \Vert \mathbf{v}^{\prime}_{\text{c},i}\right \Vert >{v_{\text{m},i}}
	\end{cases}.
\end{equation*}
It is obvious that $0<{{\kappa }_{{v_{\text{m},i}}}}\left( \mathbf{v}^{\prime}_{\text{c},i}\right)
\leq 1$. In the following, ${{\kappa }_{{v_{\text{m},i}}}}\left(\mathbf{v}^{\prime}_{\text{c},i}\right) $ will be written as ${{\kappa }_{{v_{\text{m},i}}}}$ for short. 

\textbf{Remark 1}. When a given robot is modeled as a single integrator just as (\ref{SingleIntegral}), such as multicopters, helicopters and certain types of wheeled robots equipped with omni-directional wheels \cite{Quan(2017)}, the designed velocity command $\mathbf{v}_{\text{c},i}$ can be directly applied to control the robot. When the robot model considered is more complicated, such as a second order integrator model or a nonholonomic model, additional control laws are necessary \cite{Rezende(2020),rezende2018robust}. Besides, in our previous work \cite{Quan(2021)}, we propose a \emph{filtered position model} converting a second-order model to a first-order model just as \eqref{SingleIntegral}.

\subsection{Curve Virtual Tube Model}

Before the curve virtual tube is introduced, some necessary concepts are first proposed.

\begin{figure}[h]
	\begin{centering}
		\includegraphics[scale=1]{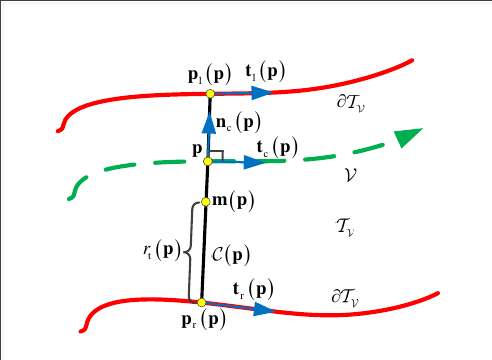}
		\par \end{centering}
	\caption{Relative concepts about the curve virtual tube.}
	\label{Curvetube}
\end{figure}


\begin{enumerate}
	\item[(i)] \textbf{Generating Curve}. A curve in $\mathbb{R}^{2}$ is \emph{simple} if it does not cross itself. When a curve starts and ends at the same point, it is a \emph{closed} curve or \emph{loop} \cite[p. 915]{Thomas(2009)}. In this
	paper, we only consider a simple and not closed \emph{generating curve} $\mathcal{V} \subset \mathbb{R}^{2}$, which starts at $\mathbf{p}_\text{s} \in \mathcal{V}$ and ends at $\mathbf{p}_\text{f} \in \mathcal{V}$. As shown in Figure \ref{Curvetube}, if there exists $\mathbf{p}\in \mathcal{V}$, define $\mathbf{t}_{\text{c}}\left( 
	\mathbf{p}\right) \in \mathbb{R}^{2}$ to be the \emph{unit tangent vector} pointing in the forward direction, namely the moving direction. Similarly, $\mathbf{n}_{\text{c}}\left( \mathbf{p}\right) \in \mathbb{R}^{2}$ is
	the \emph{unit normal vector} directing anti-clockwise or left of the tangent direction. Then it is obtained that
	\begin{equation*}
		\mathbf{t}_{\text{c}}^{\text{T}}\left( \mathbf{p}\right) \mathbf{n}_{\text{c}%
		}\left( \mathbf{p}\right) \equiv 0.
	\end{equation*}
	In the following, the generating curve $\mathcal{V}$ is considered as predefined by planning, namely this paper does not cover how to design $\mathcal{V}$.
	
	\item[(ii)] \textbf{Cross Section}. For any $\mathbf{p}\in \mathcal{V},$ a \emph{cross section} passing $\mathbf{p}$ is defined as
	\begin{align*}
		\mathcal{C}\left(\mathbf{p}\right) =&\left \{  \mathbf{x}\in {{%
				\mathbb{R}}^{2}}: \mathbf{x}={{\mathbf{p}}}+\lambda
		\left( \mathbf{p}\right) \mathbf{n}_{\text{c}}\left( \mathbf{p}\right) ,\right. \\
		&\left. \lambda _{\text{l}}\left( \mathbf{p}\right) \leq
		\lambda \left( \mathbf{p}\right) \leq \lambda _{\text{r}}\left( \mathbf{p}%
		\right) ,\lambda _{\text{l}}\left( \mathbf{p}\right) ,\lambda _{\text{r}%
		}\left( \mathbf{p}\right) \in 	\mathbb{R}	\right \} .
	\end{align*}
	Here, $\mathcal{C}\left( {{\mathbf{p}}_{\text{f}}}\right)$ is called the \emph{finishing line} or \emph{finishing cross section}.
	For any point $\mathbf{p}^{\prime }\in \mathcal{C}\left( \mathbf{p}\right),$ $\mathcal{C}\left( \mathbf{p}^{\prime }\right) $ is defined as a cross section passing $\mathbf{p}^{\prime }$. It is obvious that $\mathcal{C}
	\left( \mathbf{p}^{\prime }\right) =\mathcal{C}\left( \mathbf{p}\right) .$ Besides, it has $\mathbf{t}_{\text{c}}\left( \mathbf{p}^{\prime }\right) =\mathbf{t}_{\text{c}}\left( 
	\mathbf{p}\right)$. For $\mathbf{x}_{1}\in \mathcal{C}\left( \mathbf{p}_{1}\right) $, $
	\mathbf{x}_{2}\in \mathcal{C}\left( \mathbf{p}_{2}\right)$ and $\mathbf{p}_{1},%
	\mathbf{p}_{2}\in \mathcal{V},$ if a point can move along the tangent vector 
	$\mathbf{t}_{\text{c}}\left( \mathbf{\cdot }\right) $ from $\mathbf{p}_{1}$
	to $\mathbf{p}_{2},$ then we say $\mathbf{x}_{1}$ or $\mathcal{C}\left( 
	\mathbf{p}_{1}\right) $ locates at \emph{back} of $\mathcal{%
		C}\left( \mathbf{p}_{2}\right) .$ In other words, $\mathbf{x}_{2}$ or $\mathcal{C}%
	\left( \mathbf{p}_{2}\right) $ locates at \emph{front} of $
	\mathcal{C}\left( \mathbf{p}_{1}\right) .$
	Two endpoints of $\mathcal{C}\left(\mathbf{p}\right)$ are defined as
	\begin{align*}
		\mathbf{p}_{\text{l}}\left( \mathbf{p}\right) & \triangleq {{\mathbf{p}}}%
		+\lambda _{\text{l}}\left( \mathbf{p}\right) \mathbf{n}_{\text{c}}\left( 
		\mathbf{p}\right),  \\
		\mathbf{p}_{\text{r}}\left( \mathbf{p}\right) & \triangleq {{\mathbf{p}}}%
		+\lambda _{\text{r}}\left( \mathbf{p}\right) \mathbf{n}_{\text{c}}\left( 
		\mathbf{p}\right) .
	\end{align*}
	The \emph{width} of a cross section $\mathcal{C}\left( \mathbf{p}\right) $
	is expressed as 2$r_{\text{t}}\left( \mathbf{p}\right),$ which is defined as
	\begin{equation*}
		r_{\text{t}}\left( \mathbf{p}\right) \triangleq \frac{1}{2}\left \vert
		\lambda _{\text{r}}\left( \mathbf{p}\right) -\lambda _{\text{l}}\left( 
		\mathbf{p}\right) \right \vert .
	\end{equation*}
	For any $\mathbf{p}^{\prime }\in \mathcal{C}\left(\mathbf{p}\right),$ the \emph{middle point} of $\mathcal{C}\left( \mathbf{p}\right)$  is defined as 
	\begin{equation*}
		\mathbf{m}\left( \mathbf{p}^{\prime }\right)=\mathbf{m}\left( \mathbf{p}\right)   \triangleq \frac{1}{2}\left( \mathbf{p}_{\text{l}}\left( \mathbf{p}\right)	+\mathbf{p}_{\text{r}}\left( \mathbf{p}\right) \right) .
	\end{equation*}%
	
\end{enumerate}

Based on the concepts of the generating curve and the cross section, a curve virtual tube model is proposed. The length of the curve virtual tube is also defined.

\begin{enumerate}
	\item[(i)] \textbf{Curve Virtual Tube}.  The curve virtual tube $\mathcal{T}_{\mathcal{V}}$ is generated by keeping cross sections always perpendicular to the tangent vectors of the given generating curve, which is denoted by
	\begin{equation}
		\mathcal{T}_{\mathcal{V}}=\underset{\mathbf{p}\in \mathcal{V}}\cup \mathcal{C}\left( \mathbf{p}\right) .  \label{infinitecrosssection}
	\end{equation}%
	Then the \emph{tube boundary} $\partial \mathcal{T}_{\mathcal{V}}$ is expressed as
	\begin{equation*}
		\partial \mathcal{T}_{\mathcal{V}}=\left \{  \mathbf{x}\in {{%
				\mathbb{R}}^{2}}:
		\mathbf{x}=\mathbf{p}_{\text{l}}\left( \mathbf{p}\right) \cup \mathbf{p}_{\text{r}}\left( \mathbf{p}\right) ,\mathbf{p}\in \mathcal{V}\right \} .
	\end{equation*}
	It is obvious that $\partial \mathcal{T}_{\mathcal{V}}$ corresponds to two smooth boundary curves as shown in Figure \ref{Curvetube}. Similar with $\mathbf{t}_{\text{c}}\left( 
	\mathbf{p}\right)$, the vectors $\mathbf{t}_{\text{l}}\left( 
	\mathbf{p}\right)$ and $ \mathbf{t}_{\text{r}}\left( 
	\mathbf{p}\right)$ are defined to be the unit tangent vectors of two tube boundary curves.
	Besides, for any point $\mathbf{p}^{\prime }\in \mathcal{C}\left( \mathbf{p}\right),$ it has $\mathbf{t}_{\text{l}}\left( \mathbf{p}^{\prime }\right) =\mathbf{t}_{\text{l}}\left( 
	\mathbf{p}\right)$ and $\mathbf{t}_{\text{r}}\left( \mathbf{p}^{\prime }\right) =\mathbf{t}_{\text{r}}\left(\mathbf{p}\right)$.
	\begin{figure}
		\centering
		\includegraphics[width=\columnwidth]{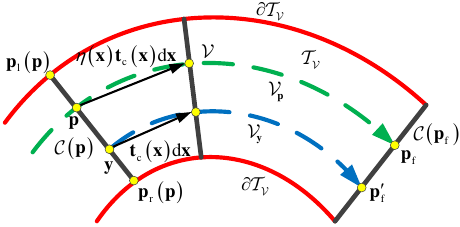}
		\caption{Brief introduction to the length and the equation \eqref{dl}.}
		\label{lineintegrallypulov1}
	\end{figure}
	
	
	\item[(ii)] \textbf{Length}. As shown in Figure \ref{lineintegrallypulov1}, given a point $
	\mathbf{y}\in \mathcal{C}\left( \mathbf{p}\right) ,$  $\mathcal{V}_{\mathbf{y}}$ is defined to be the curve from $\mathbf{y}$ to $\mathcal{C}\left( \mathbf{%
		p}_{\text{f}}\right)$ along the tangent vector $\mathbf{t}_{%
		\text{c}}\left( \mathbf{\cdot }\right) \in \mathbb{R}^{2},$ namely the arc
	from $\mathbf{y}$ to $\mathbf{p}_{\text{f}}^{\prime } \in \mathcal{C}\left( \mathbf{p}_{\text{f}}\right)$. It can be seen that the curve $\mathcal{V}_{\mathbf{y}}$ is always parallel to the generating curve $\mathcal{V}$.\footnote{The points in the same cross section have parallel tangent vectors.} Then, the length from $\mathbf{y}$ to $\mathcal{C}\left( \mathbf{p}_{\text{f}}\right) $
	is defined as the length of $\mathcal{V}_{\mathbf{p}}$ rather than $\mathcal{V}_{\mathbf{y}}$, which is denoted by $l\left( \mathbf{y}\right) .$ Here $\mathcal{V}_{\mathbf{p}}$ is a part of $\mathcal{V}$ from $\mathbf{p}$ to $\mathbf{{p}}_\text{f}$. According to this
	definition, it is obvious that points in the same cross section have the same length as $%
	\mathcal{C}\left( \mathbf{p}_{\text{f}}\right) $, namely $l\left(\mathbf{y}\right)=l\left(\mathbf{p}\right)$. If $\mathbf{y}$ locates
	at front (back) of $\mathcal{C}\left( \mathbf{p}_{%
		\text{f}}\right) ,$ the corresponding length is positive (negative). When $%
	\mathcal{C}\left( \mathbf{y}\right) =\mathcal{C}\left( \mathbf{p}_{\text{f}%
	}\right) ,$ the length is zero. 
	\begin{figure}
		\centering
		\includegraphics[width=\columnwidth]{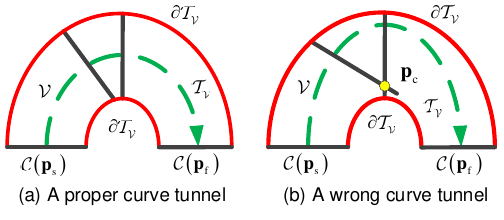}
		\caption{Proper and wrong curve virtual tubes.}
		\label{proper}
	\end{figure}
	
\end{enumerate}

Then, an assumption is made on the proposed curve virtual tube.

\textbf{Assumption 1}. For any $\mathbf{p}^{\prime }\in \mathcal{T}_{%
	\mathcal{V}},$ there exists a unique $\mathbf{p}\in \mathcal{V}$ such
that $\mathbf{p}^{\prime }\in \mathcal{C}\left( \mathbf{p}\right) .$ Besides, $\lambda _{\text{l}}\left( \mathbf{p}^{\prime }\right) $, $\lambda _{\text{r}
}\left( \mathbf{p}^{\prime }\right) $, $\mathbf{t}_{\text{c}}\left( \mathbf{p}^{\prime }\right) $, $\mathbf{t}_{\text{l}}\left( \mathbf{p}^{\prime }\right) $, $ \mathbf{t}_{\text{r}}\left( \mathbf{p}^{\prime }\right)$ are all continuous and differentiable.  

\textit{Assumption 1 }implies that, any $\mathbf{p}^{\prime }\in \mathcal{T%
}_{\mathcal{V}}$ has a unique direction to move. The generating curve and two curves of the tube boundary are also smooth. A tube satisfying \textit{Assumption 1 }is called a \emph{proper tube}. As shown in Figure \ref{proper}, the boundaries of two curve virtual
tubes are the same, but they are two
different curve virtual tubes because of different gernerating curves. The curve virtual tube shown in Figure \ref{proper}(a) is
a proper tube, while the curve virtual 
tube shown in Figure \ref{proper}(b) is not because $\mathbf{p}_{\text{c}}$ is an intersection point of
two different cross sections. This implies that there are at least two different cross sections including $\mathbf{p}_{\text{c}}$. The following proposition gives a way for any $\mathbf{p}^{\prime }\in \mathcal{T}_{\mathcal{V}}$ to get its unique moving direction.  

\textbf{Proposition 1}. Under \textit{Assumption 1}, for any $\mathbf{p}^{\prime }\in \mathcal{T}_{\mathcal{V}}$, $\mathbf{p}\in \mathcal{V}$, if and only if 
\begin{equation*}
	\left( \mathbf{p}^{\prime }-\mathbf{p}\right) ^{\text{T}}\mathbf{t}_{
		\text{c}}\left( \mathbf{p}\right) =0,
\end{equation*}%
then $\mathbf{p}^{\prime }\in \mathcal{C}\left( \mathbf{p}\right) .$

\textit{Proof}. It is easy to prove by following the definition of $\mathcal{%
	T}_{\mathcal{V}}$ and \textit{Assumption 1. }$\square $

\subsection{Two Areas around a Robot}

Similarly to our previous work \cite{Quan(2021)}, \cite{quan2021far}, two types of circular areas around a robot used for avoidance control, namely safety area and avoidance area, are introduced. 
\begin{figure}[h]
	\begin{centering}
		\includegraphics[scale=1]{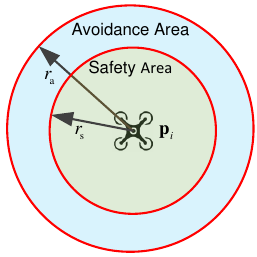}
		\par \end{centering}
	\caption{Safety area and avoidance area.}
	\label{Threeaera1}
\end{figure}
As shown in Figure \ref{Threeaera1}, at the time $t>0$, the \emph{safety area} $\mathcal{S}_{i}$ of the $i$th robot is defined as
\begin{equation*}
	\mathcal{S}_{i}\left(t\right)=\left \{ \mathbf{x}\in {{\mathbb{R}}^{2}}: \left \Vert \mathbf{x}-\mathbf{p}_{i}\left(t\right)\right \Vert \leq r_{\text{s}} \right \} ,
\end{equation*}
where $r_{\text{s}}>0$ is the \emph{safety radius}. For all robots, no \emph{conflict} with each other implies that $\mathcal{S}_{i}\cap \mathcal{S}_{j}=\varnothing$,
namely $\left \Vert \mathbf{p}_{i}-\mathbf{p}_{j}\right \Vert >2r_{\text{s}}$,
where $i,j=1,\cdots ,M,i\neq j$. Besides, the \emph{avoidance area} is defined for starting the avoidance control. As shown in Figure \ref{Threeaera1}, at the time $t>0$, the avoidance area $\mathcal{A}_{i}$ of the $i$th robot is defined as 
\begin{equation*}
	\mathcal{A}_{i}\left(t\right)=\left \{ \mathbf{x}\in {{\mathbb{R}}^{2}}:\left
	\Vert \mathbf{x}- \mathbf{p}_{i}\left(t\right)\right \Vert \leq r_{\text{a}}
	\right \}  ,
\end{equation*}
where $r_{\text{a}}>0$ is the \emph{avoidance radius}. For collision avoidance with a pair of robots, if there exists $\mathcal{A}_{i}\cap \mathcal{S}_{j}\neq \varnothing$ and $\mathcal{A}_{j}\cap \mathcal{S}_{i}\neq \varnothing$,
namely $\left \Vert \mathbf{p}_{i}-\mathbf{p}_{j}\right \Vert \leq r_{\text{a}}+r_{\text{s}}$,
then the $i$th and $j$th robots should avoid each other. The set $\mathcal{N}_{\text{m},i}$ is defined as the collection of all mark numbers of other robots whose safety areas have intersection with the avoidance area of the $i$th robot, namely 
\begin{equation*}
	\mathcal{N}_{\text{m},i}=\left \{ j: \mathcal{A}_{i} \cap \mathcal{S}_{j}
	\neq \varnothing\right \},
\end{equation*}
where $i,j=1,\cdots ,M,i\neq j$. Besides, when the $j$th robot or the tube boundary just enters the avoidance area $\mathcal{A}_{i}$ of the $i$th robot, it is required that there be no conflict in the beginning. Therefore, at least we set that 
$r_{\text{a}}>r_{\text{s}}$. 

\subsection{Problem Formulation}

With the descriptions above, some extra assumptions are imposed to get the main problem of this paper.

\textbf{Assumption 2}. The initial condition of the $i$th robot satisfies $\mathcal{S}_{i}\left( 0\right) \subset \mathcal{T}_{\mathcal{V}},$ and$\
l\left( \mathbf{p}_{i}\left( 0\right) \right) <0,i=1,\cdots ,M$, namely all robots with their safety areas are inside the curve virtual tube and all robots locate at back of the finishing line $\mathcal{C}\left( \mathbf{p}_{\text{f}}\right)$ in the beginning.

\textbf{Assumption 3}. The robots' initial conditions satisfy $\mathcal{S}_{i}\left( 0\right)\cap \mathcal{S}_{j}\left( 0\right)=\varnothing$, namely
$\left \Vert \mathbf{p}_{i}\left( 0\right) -\mathbf{p}_{j}\left( 0\right) \right \Vert >2r_{\text{s}}$, where $i,j=1,\cdots ,M,i\neq j$.

\textbf{Assumption 4}. Once a robot arrives at the finishing line $\mathcal{C}%
\left( {{\mathbf{p}}_{\text{f}}}\right) $, it will quit this curve virtual tube not
to affect other robots behind. Mathematically, given a constant ${\epsilon }_{\text{0}}>0,$
a robot arrives near the finishing line $\mathcal{C}\left( {{\mathbf{p}}_{\text{f}}}\right) $ if 
\begin{equation}
	l\left( \mathbf{p}_{i}\right) \geq -{\epsilon }_{\text{0}}.
	\label{arrivialairway}
\end{equation}

Based on \textit{Assumptions 1-4} above, the \emph{curve virtual tube passing through problem }is stated in the following.

\textbf{Curve virtual tube passing through problem}. Under \textit{Assumptions 1-4}, design the velocity command $\mathbf{v}_{\text{c},i}$ to guide all robots to pass through the curve virtual tube $\mathcal{T}_{\mathcal{V}}$, meanwhile
avoiding colliding with each other ($\mathcal{S}_{i}\left(t\right)\cap \mathcal{S}_{j}\left(t\right)=\varnothing$) and keeping within the tube ($
\mathcal{S}_{i}\left(t\right)\cap \partial \mathcal{T}_{\mathcal{V}}=\varnothing$), where $i,j=1,\cdots ,M$, $i\neq j$, $t>0$.

\section{Preliminaries}

\subsection{Line Integrals of Vector Fields}

In order to make this paper self-contained, the concept about line integrals
of vector fields is introduced \cite[pp. 901-911]{Thomas(2009)}.
Suppose that $f\left( x,y\right) $ is a real-valued function. We wish to
integrate this function over a curve $\mathcal{V}$ lying within the domain of $f$, which is parameterized by $\mathbf{r}\left( \tau \right) =\left[r_{x}\left( \tau \right)\ r_{y}\left( \tau \right) \right] ^{\text{T}}$, $a\leq \tau \leq b$. Then, the
values of $f$ along the curve are given by a composite function $f\left(
r_{x}\left( \tau \right) ,r_{y}\left( \tau \right) \right) $. We are going to
integrate this composite function with respect to the arc length from $\tau =a$ to $\tau
=b$. 

\begin{figure}[h]
	\begin{centering}
			\includegraphics[scale=1]{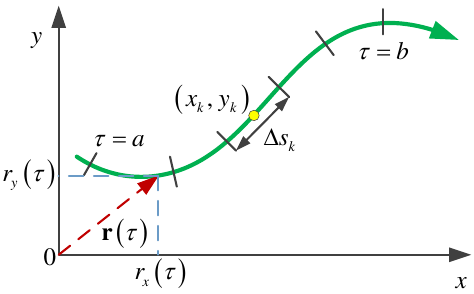}
			\par \end{centering}
	\caption{The curve $\mathbf{r}(\tau)$ is partitioned into small arcs from $\protect%
			\tau =a$ to $\protect \tau =b$. The length of a typical subarc is $\Delta
			s_{k}.$}
	\label{partitioned}
\end{figure}


As shown in Figure \ref{partitioned}, given a finite number $n$, the curve $\mathcal{V}$ is partitioned into $n$ subarcs. The length of the typical $k$th subarc is $\Delta s_{k}$. In each subarc, we choose a point $(x_{k},y_{k})$ and a sum is formed as 
\begin{equation*}
	S_{n}=\underset{k=1}{\overset{n}{\sum }}f\left( x_{k},y_{k}\right) \Delta
	s_{k},
\end{equation*}%
which is similar to a Riemann sum. Depending on how we partition the curve $%
\mathcal{V}$ and pick $(x_{k},y_{k})$ in the $k$th subarc, we may get
different values for $S_{n}$. If $f$ is continuous and the functions $%
r_{x}\left( \tau \right) $ and $r_{y}\left( \tau \right) $ have continuous
first derivatives, these sums approach a limit as $n$ increases, meanwhile the
lengths $\Delta s_{k}$ approach zero. This limit gives the following definition, similar to that for a single integral. In the definition, it is 
assumed that the partition satisfies $\Delta s_{k}\rightarrow 0$ as $%
n\rightarrow \infty .$

\textbf{Definition 1}. If $f$ is defined on a curve $\mathcal{V}$ given
parametrically by $\mathbf{r}\left( \tau \right),\ a\leq \tau \leq b$,
then the line integral of $f$ over $\mathcal{V}$ is expressed as $\int_{\mathcal{V}}f\left( x,y\right) \text{d}s.$

The parametrization $\mathbf{r}\left( \tau \right) $ defines a direction
along $\mathcal{V}$, which is called as the forward direction. At each point along
the curve $\mathcal{V}$, the tangent vector 
\begin{equation}
	\mathbf{t}_{\text{c}}=\frac{\text{d}\mathbf{r}}{\text{d}s} \label{tcdrds}
\end{equation}%
is a unit vector tangent pointing in the \emph{forward direction}.
Intuitively, the line integral of the vector field $\mathbf{f}$ is the line
integral of the scalar tangential component of $\mathbf{f}$ along $\mathcal{V%
}$. This tangential component is given by%
\begin{equation}
	\mathbf{f}^{\text{T}}\mathbf{t}_{\text{c}}=\frac{\mathbf{f}^{\text{T}}\text{d%
		}\mathbf{r}}{\text{d}s}. \label{ftc}
\end{equation}%
Then, according to \textit{Definition 1}, the line integral of $
\mathbf{f}$ along $\mathcal{V}$ is shown as
\begin{equation*}
	\int_{\mathcal{V}}\mathbf{f}^{\text{T}}\mathbf{t}_{\text{c}}\text{d}s=\int_{%
		\mathcal{V}}\frac{\mathbf{f}^{\text{T}}\text{d}\mathbf{r}}{\text{d}s}\text{d}%
	s=\int_{\mathcal{V}}\mathbf{f}^{\text{T}}\text{d}\mathbf{r}.
\end{equation*}%
Evaluate the line integral with respect to the parameter $a\leq \tau \leq b$ and it is obtained that
\begin{equation}
	\int_{\mathcal{V}}\mathbf{f}^{\text{T}}\text{d}\mathbf{r=}\int_{a}^{b}%
	\mathbf{f}\left( \mathbf{r}\left( \tau \right) \right) ^{\text{T}}\frac{%
		\text{d}\mathbf{r}\left( \tau \right)}{\text{d}\tau }\text{d}\tau .  \label{Evaluate}
\end{equation}

\subsection{Line Integral Lyapunov Function}
Given a specific point $\mathbf{p}\in {\mathcal{V}},$ its length to $\mathcal{C}%
\left( \mathbf{p}_{\text{f}}\right) $ is expressed as%
\begin{equation*}
	l\left( \mathbf{p}\right) =\int_{\mathcal{V}_{\mathbf{p}}}\text{d}l\left( 
	\mathbf{x}\right) . 
\end{equation*}
According to \eqref{tcdrds} and  \eqref{ftc}, there exists 
\begin{equation*}
	\text{d}l\left( \mathbf{x}\right) = \mathbf{t}_{\text{c}}^{\text{T}}\left( \mathbf{x}\right) \text{d}\mathbf{x},
\end{equation*}
where $\text{d}l\left( \mathbf{x}\right) $ is an increased length at $\mathbf{x}\in {\mathcal{V}%
}$. Then given a general point $\mathbf{y}\in \mathcal{T}_{\mathcal{V}},$ its length to $\mathcal{C}%
\left( \mathbf{p}_{\text{f}}\right) $ is similarly defined as
\begin{equation*}
	l\left( \mathbf{y}\right) =\int_{\mathcal{V}_{\mathbf{y}}}\text{d}l\left( 
	\mathbf{x}\right),   \label{lp}
\end{equation*}%
where $\text{d}l\left( \mathbf{x}\right) $ is an increased length at $\mathbf{x}\in \mathcal{T}_{\mathcal{V}}$. However, as shown in Figure \ref{lineintegrallypulov1}, the increased length of $\mathcal{V}_{\mathbf{y}}$ projected on $\mathcal{V}_{\mathbf{p}}$ has to multiply by $\eta \left( \mathbf{x}\right) $, which is a scaling
factor satisfying $0<\eta _{\min }<\eta \left( \mathbf{x}\right) <\eta
_{\max }<\infty $. Hence, it is obtained that 
\begin{equation}
	\text{d}l\left( \mathbf{x}\right) =\eta \left( \mathbf{x}\right) \mathbf{t}_{%
		\text{c}}^{\text{T}}\left( \mathbf{x}\right) \text{d}\mathbf{x}.  \label{dl}
\end{equation}%
It should be noted that when $\mathcal{V}_{\mathbf{y}}$ is shorter than $\mathcal{V}_{\mathbf{p}}$, there exits $\eta \left( \mathbf{x}\right) <1$, otherwise $\eta \left( \mathbf{x}\right) \geq 1$.

Then, a line integral Lyapunov function for vectors is defined as%
\begin{equation}
	V_{\text{li}}\left( \mathbf{y}\right) =\int_{\mathcal{V}_{\mathbf{y}}}\text{sat}\left( k_{1}l\left( \mathbf{x}\right) \eta \left( \mathbf{x}%
	\right) \mathbf{t}_{\text{c}}\left( \mathbf{x}\right) ,{v_{\text{m},i}}\right)
	^{\text{T}}\text{d}\mathbf{x},  \label{Vli0}
\end{equation}
where $k_{1}>0$. In the following lemma, its properties will be shown.

\textbf{Lemma 1}. Suppose that the line integral Lyapunov function $V_{\text{%
		li}}\left( \mathbf{y}\right)$ is defined as (\ref{Vli0}). Then it is obtained that (i) $V_{\text{li}}\left( \mathbf{y}\right) >0$ if $\left \vert l\left( \mathbf{y}\right) \right \vert \neq 0$; (ii) if $\left \Vert \mathbf{y}\right \Vert \rightarrow \infty ,$ then $V_{\text{li}}\left( \mathbf{y}\right) \rightarrow \infty ;$ (iii) if $V_{\text{%
		li}}\left( \mathbf{y}\right) $ is bounded, then $\left \Vert \mathbf{y}\right \Vert $ is bounded.

\emph{Proof}. See \emph{Appendix A}. $\square$

\subsection{Single Panel Method}

Panel methods are widely used in aerodynamics calculations to obtain the solution to the potential flow problem around arbitrarily shaped bodies \cite{rimon1990exact}. Here the single panel method is used to represent the repulsive potential field of the tube boundary. Assume that there is a line segment $\left[\mathbf{a}+g_1\mathbf{t},\mathbf{a}+g_2\mathbf{t}\right]$, $\left \Vert \mathbf{t} \right \Vert=1$, $g_1,g_2 \in \mathbb{R}$, the potential at any point $\mathbf{p}$ induced by the sources contained within a small element $\text{d}x$ is shown as
\begin{equation*}
	\text{d}\phi=\ln\left(\left \Vert \mathbf{p}-\left(\mathbf{a}+x\mathbf{t}\right)\right \Vert-d\right)\text{d}x,
\end{equation*}
where $d\geq0$ is a threshold distance. The induced repulsive potential function by the whole panel $\left[\mathbf{a}+g_1\mathbf{t},\mathbf{a}+g_2\mathbf{t}\right]$ is expressed as 
\begin{align}
	\phi\left(\mathbf{p},\mathbf{a},\mathbf{t},g_1,g_2,d\right) =\int_{g_1}^{g_2}\ln\left(\left \Vert \mathbf{p}-\left(\mathbf{a}+x\mathbf{t}\right)\right \Vert-d\right)\text{d}x. \label{PanelPF}
\end{align}

\begin{figure}[h]
	\begin{centering}
			\includegraphics[scale=1.4]{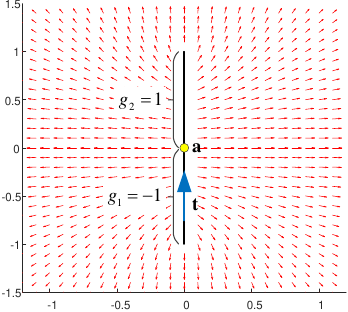}
			\par \end{centering}
	\caption{Vector field of a single panel \cite{rimon1990exact}.}
	\label{PanelVF}
\end{figure}


Given $\mathbf{a}=\left[0\ 0\right]^{\text{T}}$, $\mathbf{t}=\left[0\ 1\right]^{\text{T}}$, $g_1=-1$, $g_2=1$, $d=0$, the corresponding negative gradient vector field $-\partial \phi / \partial \mathbf{p}$ is shown in Figure \ref{PanelVF}. It can be seen that the orientation is orthogonal to the line segment $\left[\mathbf{a}+g_1\mathbf{t},\mathbf{a}+g_2\mathbf{t}\right]$ when the point $\mathbf{p}$ locates at the line $y=0$. The orientation is parallel to $\left[\mathbf{a}+g_1\mathbf{t},\mathbf{a}+g_2\mathbf{t}\right]$ when the point $\mathbf{p}$ locates at the line $x=0$. As the potential function $\phi$ is smooth and differentiable, the orientation of the vector field also changes smoothly. This phenomenon is important for the proof of no deadlock in the following. 

\subsection{Two Smooth Functions}

\begin{figure}[h]
	\begin{centering}
		\includegraphics[scale=1]{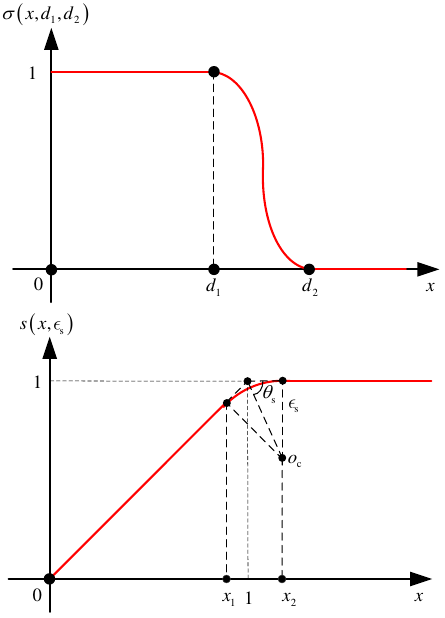}
		\par \end{centering}
	\caption{Two smooth functions. For a smooth saturation function, $\theta_\text{s}=67.5^{\circ}$.}
	\label{SmoothFunction}
\end{figure}

Two smooth functions are defined for the design of Lyapunov-like barrier functions in the following \cite{panagou2015distributed}. As shown in Figure \ref{SmoothFunction} (upper plot), the first is a second-order differentiable ``bump'' function
\begin{equation}
	\sigma \left(x,d_{1},d_{2}\right)=\left \{ 
	\begin{array}{c}
		1 \\ 
		Ax^{3}+Bx^{2}+Cx+D \\ 
		0%
	\end{array}%
	\right.
	\begin{array}{c}
		x\leq d_{1} \\ 
		d_{1}\leq x\leq d_{2} \\ 
		d_{2}\leq x%
	\end{array}
	\label{zerofunction}
\end{equation}
with $A=-2\left/\left(d_{1}-d_{2}\right)^{3}\right.,$ $B=3\left(d_{1}+d_{2}%
\right)\left/\left(d_{1}-d_{2}\right)^{3}\right.,$ $C=-6d_{1}d_{2}\left/
\left(d_{1}-d_{2}\right)^{3}\right.$,  $D=d_{2}^{2}\left(3d_{1}-d_{2}
\right)\left/\left(d_{1}-d_{2}\right)^{3}\right.$. The derivative of $\sigma \left(x,d_{1},d_{2}\right)$ with respect to $x$ is 
\begin{equation*}
	\frac{\partial\sigma \left(x,d_{1},d_{2}\right)}{\partial x}=\left \{ 
	\begin{array}{c}
		0 \\ 
		3Ax^{2}+2Bx+C \\ 
		0
	\end{array}
	\right.
	\begin{array}{c}
		x\leq d_{1} \\ 
		d_{1}\leq x\leq d_{2} \\ 
		d_{2}\leq x%
	\end{array}.
\end{equation*}

As shown in Figure \ref{SmoothFunction} (lower plot), to approximate a saturation function
\begin{equation*}
	\overline{s}\left(x\right)=\min\left(x,1\right),x\geq0,
\end{equation*}
the other is a smooth saturation function
\begin{equation*}
	s\left(x,\epsilon_{\text{s}}\right)=\left \{ 
	\begin{array}{c}
		x \\ 
		\left(1-\epsilon_{\text{s}}\right)+\sqrt{\epsilon_{\text{s}%
			}^{2}-\left(x-x_{2}\right)^{2}} \\ 
		1%
	\end{array}%
	\right.%
	\begin{array}{c}
		0\leq x\leq x_{1} \\ 
		x_{1}\leq x\leq x_{2} \\ 
		x_{2}\leq x%
	\end{array}
	\label{sat}
\end{equation*}
with $x_{2}=1+\frac{1}{\tan67.5^{\circ}}\epsilon_{\text{s}}$ and $
x_{1}=x_{2}-\sin45^{\circ}\epsilon_{\text{s}}$. Since it
is required $x_{1}>0$, one has $\epsilon_{\text{s}} \leq \frac{\tan67.5^{\circ}}{\tan67.5^{\circ}\sin45^{\circ}-1}$. For any $\epsilon_{\text{s}} \in \left[0,\frac{\tan67.5^{\circ}}{\tan67.5^{\circ}\sin45^{\circ}-1}\right]$, it is easy to see 
\begin{equation*}
	s\left(x,\epsilon_{\text{s}}\right) \leq \overline{s}\left(x\right)
\end{equation*}
and
\begin{equation*}
	\lim_{\epsilon_{\text{s}} \rightarrow 0}\sup_{x\geq 0}\left \vert \overline{s}\left(x\right)-s\left(x,\epsilon_{\text{s}}\right) \right \vert =0.
\end{equation*}
The derivative of $s\left(x,\epsilon_{\text{s}}\right)$ with respect to $x$ is
\begin{equation*}
	\frac{\partial s\left(x,\epsilon_{\text{s}}\right)}{\partial x}=\left \{ 
	\begin{array}{c}
		1 \\ 
		\frac{x_2-x}{\sqrt{\epsilon_{\text{s}}^2-\left(x-x_2\right)^2}}\\ 
		0
	\end{array}
	\right.
	\begin{array}{c}
		0\leq x\leq x_{1} \\ 
		x_{1}\leq x\leq x_{2} \\ 
		x_{2}\leq x%
	\end{array}.
\end{equation*}
For any $\epsilon_{\text{s}}>0$, we have $\sup_{x\geq 0}\left \vert \partial s\left(x,\epsilon_{\text{s}}\right) / \partial x \right \vert \leq 1$.

\section{Distributed Robotic Swarm Control for Passing through a Curve Virtual Tube}


\subsection{Line Integral Lyapunov Function for Approaching Finishing Line}


Define $\mathcal{V}_{\mathbf{p}{_{i}}}$ to be the curve from $%
\mathbf{p}{_{i}}$ to $\mathcal{C}\left( \mathbf{p}_{\text{f}}\right) $
along the tangent direction $\mathbf{t}_{\text{c}}\left( \boldsymbol{\cdot }%
\right) \in \mathbb{R}^{2}$, which is always parallel to $\mathcal{V}$. According to (\ref{Vli0}), a line integral
Lyapunov function is defined as%
\begin{equation}
	V_{\text{l},i}=\int_{\mathcal{V}_{ \mathbf{p}{_{i}}}}\text{sat}\left( k_{1}l\left( \mathbf{x}\right) \eta \left( \mathbf{x}\right) \mathbf{%
		t}_{\text{c}}\left( \mathbf{x}\right) ,{v_{\text{m},i}}\right) ^{\text{T}}
	\text{d}\mathbf{x},  \label{Vli}
\end{equation}%
where $i=1,2,\cdots ,M$. According to (\ref{Evaluate}), evaluate the line
integral with respect to the parameter $t$ and it can be obtained that 
\begin{align}
	V_{\text{l},i}=&\int_{0}^{t}\text{sat}\left( k_{1}l\left( 
	\mathbf{p}{_{i}}\right) \eta \left(  \mathbf{p}{_{i}}\right) 
	\mathbf{t}_{\text{c}}\left(  \mathbf{p}{_{i}}\right) ,{v_{\text{m},i}}%
	\right) ^{\text{T}} \dot{\mathbf{p}}{_{i}}\text{d}\tau \notag \\
	=&\int_{0}^{t}\text{sat}\left( k_{1}l\left( 
	\mathbf{p}{_{i}}\right) \eta \left(  \mathbf{p}{_{i}}\right) 
	\mathbf{t}_{\text{c}}\left(  \mathbf{p}{_{i}}\right) ,{v_{\text{m},i}}%
	\right) ^{\text{T}} {\mathbf{v}}_{\text{c},i}\text{d}\tau.
	\label{Vli1}
\end{align}
The objective of the designed velocity command is to make $V_{\text{l},i}$
approach zero. This implies that $\left \vert l\left(  \mathbf{p}{_{i}}\right)\right \vert=0$ according to \textit{Lemma 1}, namely the $i$th
robot has arrived at the finishing line $\mathcal{C}\left( {{\mathbf{p}}_{\text{f}}}%
\right) $. 

\subsection{Barrier Function for Avoiding Conflict}
Define a position error between the $i$th and $j$th robot, which is shown as
\begin{equation*}
	\tilde{\mathbf{p}}_{\text{m},ij} \triangleq \mathbf{p}_{i}-{{\mathbf{p}}_{j}}.
\end{equation*}
With the definition above, its derivative is
\begin{equation*}
	\dot{\tilde{\mathbf{p}}}_{\text{m},ij} =\mathbf{v}_{\text{c},i}-
	\mathbf{v}_{\text{c},j}  .
\end{equation*}
According to two smooth functions, the barrier function for the $i$th robot avoiding conflict with the $j$th robot is defined as
\begin{equation}
	V_{\text{m},ij}=\frac{k_{2}\sigma _{\text{m}}\left( \left \Vert \tilde{\mathbf{{p}}}{_{\text{m,}ij}}\right \Vert \right) }{\left( 1+\epsilon _{\text{%
				m}}\right) \left \Vert \tilde{\mathbf{{p}}}{_{\text{m,}ij}}\right \Vert
		-2r_{\text{s}}s\left( \frac{\left \Vert \tilde{\mathbf{{p}}}{_{\text{m,}ij}}\right \Vert }{2r_{\text{s}}},\epsilon _{\text{s}}\right) },
	\label{Vmij}
\end{equation}
where $k_2,\epsilon _{\text{m}},\epsilon _{\text{s}}>0$. Based on the definitions of the safety area and the avoidance area, the smooth function $\sigma \left( \cdot \right) $ in (\ref{zerofunction}) is defined as $\sigma _{\text{m}}\left( x\right) \triangleq \sigma \left( x,2r_{\text{s}},r_{\text{a}}+r_{\text{s}}\right)$. The function $V_{\text{m},ij}$ has the following properties.
\begin{enumerate}
	\item[(i)] $\partial V_{\text{m},ij} / \partial \left \Vert\tilde{\mathbf{p}}{_{\text{m,}ij}}\right \Vert\leq 0$ as $V_{\text{m},ij}$ is a nonincreasing function with respect to $\left \Vert\tilde{\mathbf{p}}{_{\text{m,}ij}}\right \Vert$.
	\item[(ii)] If $\left \Vert\tilde{\mathbf{p}}{_{\text{m,}ij}}\right \Vert>r_\text{s}+r_\text{a}$, namely $\mathcal{A}_{i}\cap \mathcal{S}_{j}\neq \varnothing$ and $\mathcal{A}_{j}\cap \mathcal{S}_{i}\neq \varnothing$, then $V_{\text{m},ij}=0$ and $\partial V_{\text{m},ij} / \partial \left \Vert\tilde{\mathbf{p}}{_{\text{m,}ij}}\right \Vert= 0$; if $V_{\text{m},ij}=0$, then $\left \Vert\tilde{\mathbf{p}}{_{\text{m,}ij}}\right \Vert>r_\text{s}+r_\text{a}>2r_\text{s}$.
	\item[(iii)] If $0<\left \Vert\tilde{\mathbf{p}}{_{\text{m,}ij}}\right \Vert<2r_\text{s}$, namely $\mathcal{S}_{i}\cap \mathcal{S}_{j}\neq \varnothing$ (they may not collide in practice), then there exists a sufficiently small $\epsilon_{\text{s}}>0$ such that 
	\begin{equation*}
		V_{\text{m},ij}=\frac{k_2}{\epsilon_{\text{m}}\left \Vert\tilde{\mathbf{p}}{_{\text{m,}ij}}\right \Vert}\geq \frac{k_2}{2\epsilon_{\text{m}}r_\text{s}}.
	\end{equation*}
\end{enumerate}
The objective of the designed velocity command is to make $V_{\text{m},ij}$ zero or as small as possible, which implies that $
\left \Vert \tilde{\mathbf{{p}}}{_{\text{m,}ij}}\right \Vert >2r_{\text{s}}$ according to property (ii), namely the $i$th robot will not conflict with the $j$th robot. 

\subsection{Barrier Functions for Keeping within Curve Virtual Tube}
\subsubsection{Algebraic Description for any Robot with its Safety Area inside the Curve Virtual Tube}
\begin{figure}
	\centering
	\includegraphics[width=\columnwidth]{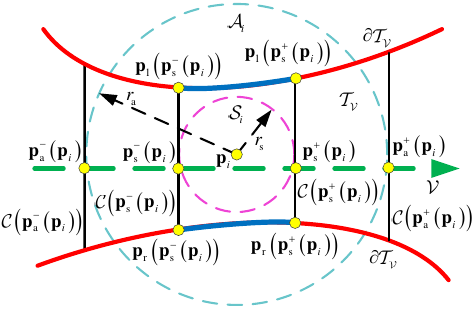}
	\caption{A visual representation about $\mathcal{S}_i\subset\mathcal{T}_{\mathcal{V}}$. This requires $\text{dist}\left(\mathbf{p}_i,	\partial \mathcal{T}_{\mathcal{V}}\cap\mathcal{C}\left(\mathbf{p}^{\prime}\right)\right)>r_\text{s}$ if $l\left(\mathbf{p}^{\prime}\right)>l\left(\mathbf{p}^{-}_\text{s}\left(\mathbf{p}_i\right)\right)$ or $l\left(\mathbf{p}^{\prime}\right)<l\left(\mathbf{p}^{+}_\text{s}\left(\mathbf{p}_i\right)\right)$, namely $\mathcal{S}_i$ should have no intersection with the curves in blue, which are parts of $\partial \mathcal{T}_{\mathcal{V}}$ from $\mathbf{p}_{\text{l}}\left( \mathbf{p}^{-}_\text{s}\left(\mathbf{p}_i\right)\right) $ to $\mathbf{p}_{\text{l}}\left( \mathbf{p}^{+}_\text{s}\left(\mathbf{p}_i\right)\right) $ and from $\mathbf{p}_{\text{r}}\left( \mathbf{p}^{-}_\text{s}\left(\mathbf{p}_i\right)\right) $ to $\mathbf{p}_{\text{r}}\left( \mathbf{p}^{+}_\text{s}\left(\mathbf{p}_i\right)\right) $.}
	\label{SATube}
\end{figure}
When the $i$th robot with its safety area $\mathcal{S}_i$ is moving inside the curve virtual tube, namely $\mathcal{S}_i\subset\mathcal{T}_{\mathcal{V}}$, $\mathcal{S}_i$ is actually confined between the crossing sections $\mathcal{C}\left(\mathbf{p}^{-}_\text{s}\left(\mathbf{p}_i\right)\right)$ and $\mathcal{C}\left(\mathbf{p}^{+}_\text{s}\left(\mathbf{p}_i\right)\right)$ as shown in Figure \ref{SATube}, where two points $\mathbf{p}^{-}_\text{s}\left(\mathbf{p}_i\right),\mathbf{p}^{+}_\text{s}\left(\mathbf{p}_i\right) \in \mathcal{V}$ with respect to $\mathbf{p}_i$ and $r_\text{s}$ are defined as
\begin{align*}
	\mathbf{p}^{-}_\text{s}\left(\mathbf{p}_i\right) &\triangleq \mathop{\text{argmin}}_{\mathbf{x}\in\mathcal{V},\text{dist}\left(\mathbf{p}_i,\mathcal{C}\left(\mathbf{x}\right)\right)<r_\text{s} }l\left(\mathbf{x}\right),\\
	\mathbf{p}^{+}_\text{s}\left(\mathbf{p}_i\right) &\triangleq \mathop{\text{argmax}}_{\mathbf{x}\in\mathcal{V},\text{dist}\left(\mathbf{p}_i,\mathcal{C}\left(\mathbf{x}\right)\right)<r_\text{s} }l\left(\mathbf{x}\right),
\end{align*}
where the function $\text{dist}\left(\cdot\right)$ is defined as the Euclidean distance. Obviously, the point $	\mathbf{p}^{-}_\text{s}\left(\mathbf{p}_i\right) $ locates at back of $\mathbf{p}_i$, and the point $	\mathbf{p}^{+}_\text{s}\left(\mathbf{p}_i\right) $ locates at front of $\mathbf{p}_i$.
For the $i$th robot, the set $\mathcal{V}_{\text{s},i}$ is defined as the collection of all points on the generating curve whose lengths are between $l\left(\mathbf{p}^{-}_\text{s}\left(\mathbf{p}_i\right)\right)$ and $l\left(\mathbf{p}^{+}_\text{s}\left(\mathbf{p}_i\right)\right)$, namely 
\begin{equation*}
	\mathcal{V}_{\text{s},i}=\left \{  \mathbf{x}\in \mathcal{V}:
	l\left(\mathbf{x}\right)\in\left(l\left(\mathbf{p}^{-}_\text{s}\left(\mathbf{p}_i\right)\right),l\left(\mathbf{p}^{+}_\text{s}\left(\mathbf{p}_i\right)\right)\right)\right \},
\end{equation*}
where $i=1,\cdots ,M$. Then an algebraic description for the $i$th robot with its safety area inside the curve virtual tube, namely $\mathcal{S}_i\subset\mathcal{T}_{\mathcal{V}} $, is expressed as
\begin{equation}
	\text{dist}\left(\mathbf{p}_i,	\partial \mathcal{T}_{\mathcal{V}}\cap\mathcal{C}\left(\mathbf{p}^{\prime}\right)\right)>r_\text{s}, \forall \mathbf{p}^{\prime} \in 	\mathcal{V}_{\text{s},i}. \label{pivsi}
\end{equation}

Besides, two points $\mathbf{p}^{-}_\text{a}\left(\mathbf{p}_i\right),\mathbf{p}^{+}_\text{a}\left(\mathbf{p}_i\right) \in \mathcal{V}$ with respect to $\mathbf{p}_i$ and $r_\text{a}$ are similarly defined as
\begin{align*}
	\mathbf{p}^{-}_\text{a}\left(\mathbf{p}_i\right) &\triangleq \mathop{\text{argmin}}_{\mathbf{x}\in\mathcal{V},\text{dist}\left(\mathbf{p}_i,\mathcal{C}\left(\mathbf{x}\right)\right)<r_\text{a} }l\left(\mathbf{x}\right),\\
	\mathbf{p}^{+}_\text{a}\left(\mathbf{p}_i\right) &\triangleq \mathop{\text{argmax}}_{\mathbf{x}\in\mathcal{V},\text{dist}\left(\mathbf{p}_i,\mathcal{C}\left(\mathbf{x}\right)\right)<r_\text{a} }l\left(\mathbf{x}\right).
\end{align*}
Obviously, the point $	\mathbf{p}^{-}_\text{a}\left(\mathbf{p}_i\right) $ locates at back of $	\mathbf{p}^{-}_\text{s}\left(\mathbf{p}_i\right) $, and the point $	\mathbf{p}^{+}_\text{a}\left(\mathbf{p}_i\right) $ locates at front of $	\mathbf{p}^{+}_\text{s}\left(\mathbf{p}_i\right) $. For the $i$th robot, the set $\mathcal{V}_{\text{a},i}$ is defined as the collection of all points on the generating curve whose lengths are between $l\left(\mathbf{p}^{-}_\text{a}\left(\mathbf{p}_i\right)\right)$ and $l\left(\mathbf{p}^{+}_\text{a}\left(\mathbf{p}_i\right)\right)$, namely 
\begin{equation*}
	\mathcal{V}_{\text{a},i}=\left \{  \mathbf{x}\in \mathcal{V}:
	l\left(\mathbf{x}\right)\in\left(l\left(\mathbf{p}^{-}_\text{a}\left(\mathbf{p}_i\right)\right),l\left(\mathbf{p}^{+}_\text{a}\left(\mathbf{p}_i\right)\right)\right)\right \},
\end{equation*}
where $i=1,\cdots ,M$. 

\begin{figure}
	\centering
	\includegraphics[width=\columnwidth]{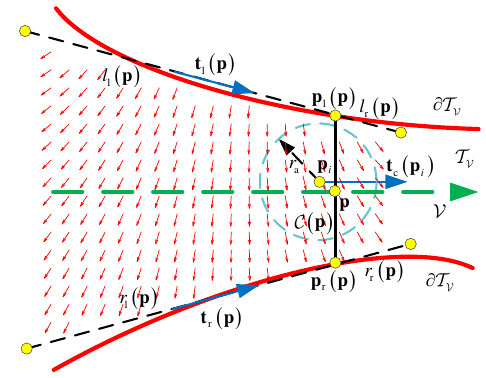}
	\caption{Relative concepts of two extended tube boundaries. Red arrows represent the negative vector field of the single panel $\left[\mathbf{p}_\text{l}\left(\mathbf{p}\right)+l_\text{r}\left(\mathbf{p}\right)\mathbf{t}_\text{l}\left(\mathbf{p}\right),\mathbf{p}_\text{l}\left(\mathbf{p}\right)+l_\text{l}\left(\mathbf{p}\right)\mathbf{t}_\text{l}\left(\mathbf{p}\right)\right]$, namely $-\left(\partial\phi_\text{l}\left(\mathbf{p}_i,\mathbf{p}\right)/\partial \mathbf{p}_i\right)^{\text{T}}$. In this figure, there exists $-\mathbf{t}_{\text{c}}^{\text{T}}\left(\mathbf{p}_i\right)\left(\partial\phi_\text{l}\left(\mathbf{p}_i,\mathbf{p}\right)/\partial \mathbf{p}_i\right)^{\text{T}}>0$}
	\label{LRVF}
\end{figure}

\subsubsection{Barrier Functions Design with Single Panel Method}
As shown in Figure \ref{LRVF}, for any $\mathbf{p} \in  \mathcal{V}$, two extended straight-line tube boundaries are defined 
\begin{align*}
	&\left[\mathbf{p}_\text{l}\left(\mathbf{p}\right)+l_\text{r}\left(\mathbf{p}\right)\mathbf{t}_\text{l}\left(\mathbf{p}\right),\mathbf{p}_\text{l}\left(\mathbf{p}\right)+l_\text{l}\left(\mathbf{p}\right)\mathbf{t}_\text{l}\left(\mathbf{p}\right)\right],\\
	&\left[\mathbf{p}_\text{r}\left(\mathbf{p}\right)+r_\text{r}\left(\mathbf{p}\right)\mathbf{t}_\text{r}\left(\mathbf{p}\right),\mathbf{p}_\text{r}\left(\mathbf{p}\right)+r_\text{l}\left(\mathbf{p}\right)\mathbf{t}_\text{r}\left(\mathbf{p}\right)\right],
\end{align*}
satisfying 
\begin{align*}
	l_\text{r}\left(\mathbf{p}\right),r_\text{r}\left(\mathbf{p}\right)&>0,\\
	l_\text{l}\left(\mathbf{p}\right),r_\text{l}\left(\mathbf{p}\right)&<0.
\end{align*}
According to the potential function of the single panel (\ref{PanelPF}), the potential fields of the two extended tube boundaries at $\mathbf{p}_i \in  \mathcal{T}_{\mathcal{V}}$ are shown as 
\begin{align*}
	&\phi\left(\mathbf{p}_i,\mathbf{p}_\text{l}\left(\mathbf{p}\right),\mathbf{t}_\text{l}\left(\mathbf{p}\right),l_\text{r}\left(\mathbf{p}\right),l_\text{l}\left(\mathbf{p}\right),r_\text{s}\right),\\
	&\phi\left(\mathbf{p}_i,\mathbf{p}_\text{r}\left(\mathbf{p}\right),\mathbf{t}_\text{r}\left(\mathbf{p}\right),r_\text{r}\left(\mathbf{p}\right),r_\text{l}\left(\mathbf{p}\right),r_\text{s}\right),
\end{align*}
which will be written as $\phi_\text{l}\left(\mathbf{p}_i,\mathbf{p}\right)$ and $\phi_\text{r}\left(\mathbf{p}_i,\mathbf{p}\right)$ for short in the following.
If there exists 
$\text{dist}\left(\mathbf{p}_i,\mathcal{C}\left(\mathbf{p}\right)\right)<r_\text{a}$, $l_\text{r}\left(\mathbf{p}\right),l_\text{l}\left(\mathbf{p}\right),r_\text{r}\left(\mathbf{p}\right),r_\text{l}\left(\mathbf{p}\right)$ must satisfy the following constraints 
\begin{align}
	-\mathbf{t}_{\text{c}}^{\text{T}}\left(\mathbf{p}_i\right)\left(\frac{\partial \phi_\text{l}\left(\mathbf{p}_i,\mathbf{p}\right)}{\partial \mathbf{p}_i }\right)^{\text{T}}&\geq0,  \label{DirL}\\
	-\mathbf{t}_{\text{c}}^{\text{T}}\left(\mathbf{p}_i\right)\left(\frac{\partial \phi_\text{r}\left(\mathbf{p}_i,\mathbf{p}\right)}{\partial \mathbf{p}_i }\right)^{\text{T}}&\geq0. \label{DirR}
\end{align}

Then, two barrier functions for achieving \eqref{pivsi}, namely the $i$th robot with its safety area keep within the tube, are defined as
\begin{align}
	V_{\text{tl},i}&=k_{3}\sum_{\mathbf{p} \in 	\mathcal{V}_{\text{a},i}}\phi_\text{l}\left(\mathbf{p}_i,\mathbf{p}\right), \label{Vtli} \\ 	V_{\text{tr},i}&=k_{3}\sum_{\mathbf{p} \in 	\mathcal{V}_{\text{a},i}}\phi_\text{r}\left(\mathbf{p}_i,\mathbf{p}\right), \label{Vtri}
\end{align}
where $k_3>0$. According to the constraints \eqref{DirL} and \eqref{DirR}, there exists
\begin{equation}
	\left\{
	\begin{aligned}
		&-\mathbf{t}_{\text{c}}^{\text{T}}\left(\mathbf{p}_i\right)\left(\frac{\partial V_{\text{tl},i}}{\partial \mathbf{p}_i }\right)^{\text{T}}\geq0\\
		&-\mathbf{t}_{\text{c}}^{\text{T}}\left(\mathbf{p}_i\right)\left(\frac{\partial V_{\text{tr},i}}{\partial \mathbf{p}_i }\right)^{\text{T}}\geq0
	\end{aligned}  \label{DirVti}
	\right..
\end{equation}
The inequalities \eqref{DirVti} mean that the angles between negative gradient directions of $V_{\text{tl},i}$, $V_{\text{tr},i}$ inside the tube and the moving direction $\mathbf{t}_{\text{c}}\left(\mathbf{p}_i\right)$ must keep smaller than $90^{\circ}$, which plays a crucial role in the stability proof. The objective of the designed velocity command is to make $V_{\text{tl},i}$, $V_{\text{tr},i}$ as small as possible, which implies that the $i$th robot will keep within the tube.

\textbf{Remark 2}. In \eqref{Vtli} and \eqref{Vtri}, the reason for applying $\mathcal{V}_{\text{a},i}$ rather than $\mathcal{V}_{\text{s},i}$ is to prevent the robot from moving outside the curve virtual tube as there may exist unpredictable uncertainties out of the assumptions we make.


\subsection{Swarm Controller Design}

Let $\mathbf{p}$ be the collection $\left(\mathbf{p}_{1},\cdots ,\mathbf{p}_{M}\right)$. The velocity command of the $i$th robot is designed as   
\begin{equation}
	\mathbf{v}_{\text{c},i}=\mathbf{v}\left( \mathcal{T}_{\mathcal{V}},\mathbf{p}_{i},\mathbf{p},r_\text{s}\right), \label{control_highway_dis}
\end{equation}%
where%
\begin{align*}
	&\mathbf{v}\left( \mathcal{T}_{\mathcal{V}},\mathbf{p}_{i},\mathbf{p},r_\text{s}
	\right) \triangleq -\text{sat}\left( \underset{\text{Line Approaching}}{%
		\underbrace{\text{sat}\left( k_{1}l\left( \mathbf{p}_i\right)
			\eta \left( \mathbf{p}_{i}\right) \mathbf{t}_{\text{c}}\left(\mathbf{p}_{i}\right) ,{v_{\text{m},i}}\right) }}\right. \notag \\
	&\left.+{\underset{\text{Robot Avoidance}}{%
			\underbrace{\underset{j\in \mathcal{N}_{\text{m},i}}{\overset{}{\sum }}%
				-b_{ij}\tilde{\mathbf{p}}_{\text{m},ij}}}}+\underset{\text{Virtual Tube
			Keeping}}{\underbrace{ \left(\frac{\partial V_{\text{tl},i}}{\partial \mathbf{p}_i }\right)^{\text{T}}+\left(\frac{\partial V_{\text{tr},i}}{\partial \mathbf{p}_i }\right)^{\text{T}}}},{
		v_{\text{m},i}}\right)  
\end{align*}%
with \footnote{$b_{ij}>0$ according to the property (i) of $V_{\text{m},ij}$.}
\begin{equation*}
	b_{ij} =-\frac{\partial V_{\text{m},ij}}{\partial \left \Vert \tilde{\mathbf{{p}}}{_{\text{m,}ij}}\right \Vert }\frac{1}{\left \Vert \tilde{\mathbf{{p}}}{_{\text{m,}ij}}\right \Vert } .
\end{equation*}
This is a distributed control form. Unlike the formation control, neighboring robots’ IDs of a robot are not required. With some active detection devices, such as cameras or radars, the proposed controller can work autonomously without wireless communication and other robots' IDs. 

\textbf{Remark 3}. It is noticed that the velocity command \eqref{control_highway_dis} is
saturated, whose norm cannot exceed $v_{\text{m},i}$. If the case such as $\left \Vert \tilde{\mathbf{{p}}}{_{\text{m,}ij_1}}\right \Vert <2r_\text{s}$ happens in practice due to unpredictable uncertainties out of the assumptions we make, this may not imply that the $i$th robot has collided the $j_1$th UAV physically. In this case, the velocity command \eqref{control_highway_dis}  degenerates to be
\begin{align*}
	\mathbf{v}_{\text{c},i}&=-\text{sat}\left(\text{sat}\left( k_{1}l\left( \mathbf{p}_i\right)
	\eta \left( \mathbf{p}_{i}\right) \mathbf{t}_{\text{c}}\left(\mathbf{p}_{i}\right) ,{v_{\text{m},i}}\right)\right.\\
	&\left.-\sum_{j=1,j\neq i,j_1}^{M}b_{ij}\tilde{\mathbf{p}}_{\text{m},ij}-b_{ij_1}\tilde{\mathbf{p}}_{\text{m},ij_1}\right.\\
	&\left.+\left(\frac{\partial V_{\text{tl},i}}{\partial \mathbf{p}_i }\right)^{\text{T}}+\left(\frac{\partial V_{\text{tr},i}}{\partial \mathbf{p}_i }\right)^{\text{T}},{v_{\text{m},i}}\right)
\end{align*}
with
\begin{equation*}
	b_{ij_1} \approx \frac{k_2}{\epsilon_\text{m}\left \Vert \tilde{\mathbf{{p}}}{_{\text{m,}ij_1}}\right \Vert^3}.
\end{equation*}
Since $\epsilon_\text{m}$ is chosen to be sufficiently small, the term $b_{ij_1}\tilde{\mathbf{p}}_{\text{m},ij_1}$ will dominate\footnote{Furthermore, we assume that the $i$th robot does not conflict with others except for the $j_1$th robot, or not very close to the boundary of the curve virtual tube.} so that the velocity
command $\mathbf{v}_{\text{c},i}$ becomes
\begin{align*}
	\mathbf{v}_{\text{c},i}\approx\text{sat}\left(\frac{k_2}{\epsilon_\text{m}\left \Vert \tilde{\mathbf{{p}}}{_{\text{m,}ij_1}}\right \Vert^3}\tilde{\mathbf{{p}}}{_{\text{m,}ij_1}},{v_{\text{m},i}}\right).
\end{align*}
This implies that $\left \Vert \tilde{\mathbf{{p}}}{_{\text{m,}ij_1}}\right \Vert$ will be increased
very fast so that the $i$th robot can keep away from the $j_1$th robot immediately.

\subsection{Stability Analysis}

In order to investigate the stability of the proposed controller for the curve virtual tube passing through problem, a function is defined as
follows%
\begin{equation*}
	{V}=\underset{i=1}{\overset{M}{\sum }}\left( V_{\text{l},i}+\frac{1}{2}%
	\underset{j=1,j\neq i}{\overset{M}{\sum }}V_{\text{m},ij}+V_{\text{tl},i}+V_{\text{tr},i}\right) ,
\end{equation*}
where $V_{\text{l},i},V_{\text{m},ij}$, $V_{\text{tl},i}$, $V_{\text{tr},i}$ are defined in (\ref{Vli}), (\ref{Vmij}), (\ref{Vtli}), (\ref{Vtri}) respectively. The derivative of ${V}$ is shown as 
\begin{align*}
	&{\dot{V}} =\sum_{i=1}^{M}\left(\text{sat}
	\left( k_{1}l\left( \mathbf{p}{_{i}}\right) \eta \left( \mathbf{p}{_{i}}\right) \mathbf{t}_{\text{c}}\left( \mathbf{p}{_{i}}%
	\right) ,{v_{\text{m},i}}\right) ^{\text{T}}\mathbf{v}_{\text{c},i}\right. \notag \\
	&\left.-\frac{{1}}{2}\sum_{j=1,j\neq i}^{M}b_{ij}\tilde{\mathbf{p}}
	_{\text{m,}ij}^{\text{T}}\left( \mathbf{v}_{\text{c},i}-\mathbf{v}_{\text{c},j}\right) +\frac{\partial V_{\text{tl},i}}{\partial \mathbf{p}_i } \mathbf{v}_{\text{c},i}+\frac{\partial V_{\text{tr},i}}{\partial \mathbf{p}_i } \mathbf{v}_{\text{c},i}\right) \\
	&=\sum_{i=1}^{M}\left( \text{sat}\left(
	k_{1}l\left( \mathbf{p}{_{i}}\right) \eta \left( \mathbf{p}{_{i}}\right) \mathbf{t}_{\text{c}}\left(\mathbf{p}{_{i}}\right) ,{v_{\text{m},i}}\right)-\sum_{j\in \mathcal{N}_{\text{m},i}}b_{ij}
	\tilde{\mathbf{p}}_{\text{m,}ij}\right. \notag \\
	&\left. +\left(\frac{\partial V_{\text{tl},i}}{\partial \mathbf{p}_i }\right)^{\text{T}}+\left(\frac{\partial V_{\text{tr},i}}{\partial \mathbf{p}_i }\right)^{\text{T}}\right) ^{\text{T}}\mathbf{v}_{\text{c},i},
\end{align*}%
where 
\begin{equation*}
	\underset{j=1,j\neq i}{\overset{M}{\sum }}b_{ij}\tilde{\mathbf{p}}
	_{\text{m,}ij}=\underset{j\in \mathcal{N}_{\text{m},i}}{\overset{}{\sum }}%
	b_{ij}\tilde{\mathbf{p}}
	_{\text{m,}ij}.
\end{equation*}%
By using the velocity command  (\ref{control_highway_dis}) for all robots, ${\dot{V}}$ satisfies
\begin{equation}
	{\dot{V}}\leq 0.  \label{dV2}
\end{equation}

\textbf{Remark 4}. In fact, for approaching the finishing line, a straightforward and intuitive way is to define a Lyapunov function as%
\begin{equation}
	V_{\text{l},i}^{\prime}=\frac{1}{2}k_{1}l^{2}\left( \mathbf{p}{_{i}}\right) ,
	\label{Vliold}
\end{equation}%
whose derivative is shown as
\begin{equation*}
	\dot{V}_{\text{l},i}^{\prime}=k_{1}l\left(  \mathbf{p}{_{i}}\right) \eta \left( 
	\mathbf{p}{_{i}}\right) \mathbf{t}_{\text{c}}^{\text{T}}\left( 
	\mathbf{p}{_{i}}\right) \mathbf{v}_{\text{c},i}.
\end{equation*}%
By replacing $V_{\text{l},i}$ in (\ref{Vli}) with $V_{\text{l},i}^{\prime}$ in {(\ref{Vliold})}, $\dot{V}$ is converted as 
\begin{align*}
	{\dot{V}}&=\underset{i=1}{\overset{M}{\sum }}\left(k_{1}l\left(  \mathbf{p}{_{i}}\right) \eta \left(  \mathbf{p}{_{i}}\right) \mathbf{t}_{
		\text{c}}\left( \mathbf{p}{_{i}}\right) -\sum_{j\in \mathcal{N}_{\text{m},i}}b_{ij}\tilde{\mathbf{p}}_{\text{m,}ij}\right.\\
	&\left.+\left(\frac{\partial V_{\text{tl},i}}{\partial \mathbf{p}_i }\right)^{\text{T}}+\left(\frac{\partial V_{\text{tr},i}}{\partial \mathbf{p}_i }\right)^{\text{T}}\right) ^{\text{T}}\mathbf{v}_{\text{c},i}.
\end{align*}%
In this way, we \textbf{CANNOT} introduce a \emph{saturation} on the finishing line
approaching term to the final controller. For such a purpose, the definition of the line
integral Lyapunov function $V_{\text{l},i}$ in (\ref{Vli}) is necessary.

Before the main result in this subsection is introduced, an important lemma is needed.


\textbf{Lemma 2}. Under \textit{Assumptions 1-4}, suppose that (i) the velocity command for the $i$th robot is designed as (\ref{control_highway_dis}); (ii) the curve virtual tube $\mathcal{\mathcal{T}_{\mathcal{V}}}$ is wide enough for at least one robot to pass through. Then there exist sufficiently small $\epsilon _{\text{m}},\epsilon _{\text{s}}>0$ in $b_{ij}$ such that $\mathcal{S}_{i}\left(t\right)\cap \mathcal{S}_{j}\left(t\right)=\varnothing,$ $\mathcal{S}_{i}\left(t\right)\cap \partial \mathcal{T}_{\mathcal{V}}=\varnothing$, $t\in \lbrack 0,\infty )$ for all ${{\mathbf{p}}_{i}(0)}$, $i,j=1,\cdots ,M$ and $i \neq j$.

\emph{Proof}. See \emph{Appendix B}. $\square$

With \emph{Lemmas 1-2} available, the main result is stated as follows.

\textbf{Theorem 1}. Under \textit{Assumptions 1-4}, suppose that (i) the velocity command for the $i$th robot is designed as (\ref{control_highway_dis}); (ii) given ${\epsilon }_{\text{0}}>0{,}$ if (\ref{arrivialairway}) is satisfied, then $b_{ij}\equiv 0$, $\left(\partial V_{\text{tl},i}/\partial \mathbf{p}_i\right)^{\text{T}}\equiv \mathbf{0}$, $\left(\partial V_{\text{tr},i}/\partial \mathbf{p}_i\right)^{\text{T}}\equiv \mathbf{0}$, which implies that the $i$th robot is removed from the curve virtual tube $\mathcal{\mathcal{T}_{\mathcal{V}}}$ mathematically; (iii) the curve virtual tube $\mathcal{\mathcal{T}_{\mathcal{V}}}$ is wide enough for at least one robot to pass through. Then, there exist sufficiently small $\epsilon _{\text{m}},\epsilon_{\text{s}}>0$ in $b_{ij}$ and $t_{1}>0 $ such that all robots can satisfy (\ref{arrivialairway}) at $t\geq t_{1},$ meanwhile guaranteeing $\mathcal{S}_{i}\left(t\right)\cap \mathcal{S}_{j}\left(t\right)=\varnothing,$ $\mathcal{S}_{i}\left(t\right)\cap \partial \mathcal{T}_{\mathcal{V}}=\varnothing$, $t\in \lbrack 0,\infty )$ for all ${{\mathbf{p}}_{i}(0)}$, $i,j=1,\cdots ,M$ and $i \neq j$.

\emph{Proof}. See \emph{Appendix C}. $\square$

\subsection{Modified Swarm Controller Design}
However, the distributed controller (\ref{control_highway_dis}) has two apparent imperfections in use. 
\begin{enumerate}
	\item[(i)]  The first problem is that any robot can approach the finishing line but its speed will slow down to zero. The reason is that $l\left( \mathbf{p}{_{i}}\right) =0$ when $\mathbf{p}_{i}$ locates at $\mathcal{C}\left( {{\mathbf{p}}_{\text{f}}}\right)$.
	\item[(ii)]  The second problem is that the values of $l_\text{r}\left(\mathbf{p}\right),l_\text{l}\left(\mathbf{p}\right),r_\text{r}\left(\mathbf{p}\right),r_\text{l}\left(\mathbf{p}\right)$ for all $\mathbf{p} \in  \mathcal{V}$ are difficult to obtain. The specific mathematical forms of $\partial V_{\text{tl},i}/\partial \mathbf{p}_i$ and $\partial V_{\text{tr},i}/\partial \mathbf{p}_i$ are also very complicated and inconvenient for practical use.
\end{enumerate}

\subsubsection{Line Approaching Term with a Constant Speed }
\begin{figure}[h]
	\begin{centering}
		\includegraphics[scale=1]{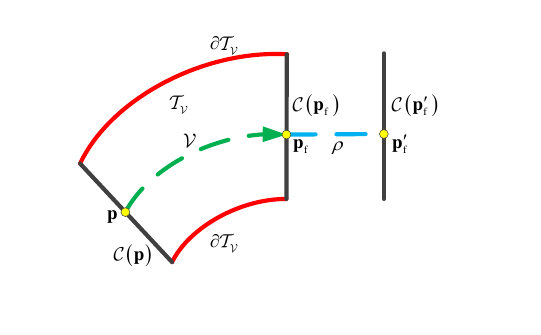}
		\par \end{centering}
	\caption{Modified Finishing Line.}
	\label{NewFinishingLine}
\end{figure}

To solve the first problem, a modified finishing line $\mathcal{C}\left( {{\mathbf{p}}}_{\text{{f}}}^{\prime }\right) $ is defined as shown in Figure {\ref{NewFinishingLine}} with a length 
\begin{equation*}
	l^{\prime}\left( \mathbf{p}{_{i}}\right)=l\left( \mathbf{p}{_{i}}\right)-\rho,
\end{equation*}
where $\rho\geq v_{\text{m},i}/\eta_{\min }$. In this case,  the line approaching term becomes
\begin{align*}
	&-\text{sat}\left( k_{1}l^{\prime}\left( \mathbf{p}_i\right)
	\eta \left( \mathbf{p}_{i}\right) \mathbf{t}_{\text{c}}\left(\mathbf{p}_{i}\right) ,{v_{\text{m},i}}\right)\\
	=&-\text{sat}\left( k_{1}\left(l\left( \mathbf{p}{_{i}}\right)-\rho\right)
	\eta \left( \mathbf{p}_{i}\right) \mathbf{t}_{\text{c}}\left(\mathbf{p}_{i}\right) ,{v_{\text{m},i}}\right)\\
	=&v_{\text{m},i}\mathbf{t}_{\text{c}}\left( \mathbf{p}_{i}\right).
\end{align*}

\subsubsection{Unified Barrier Function for Keeping within Curve Virtual Tube}
\begin{figure}[h]
	\begin{centering}
		\includegraphics[scale=1.2]{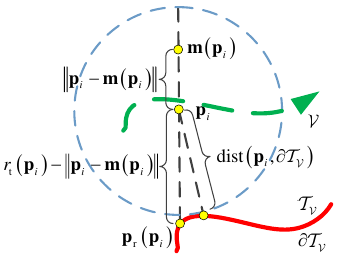}
		\par \end{centering}
	\caption{The reason for proposing the modified safety radius..}
	\label{Threeaera2}
\end{figure}

To solve the second problem, a unified barrier function is introduced to imitate the performance of  $ V_{\text{tl},i}$ and $ V_{\text{tr},i}$.
An Euclidean distance error is defined between the $i$th robot and the tube boundary confined in $\mathcal{C}\left( \mathbf{p}_{i}\right)$, which is shown as 
\begin{equation*}
	d{_{\text{t,}i}} \triangleq r_{\text{t}}\left( \mathbf{p}_{i}\right) - \left \Vert \mathbf{p}_{i}-\mathbf{m}\left( \mathbf{p}_i\right) \right \Vert.
\end{equation*}
The derivative of this error is shown as 
\begin{equation*}
	\dot{d}{_{\text{t,}i}}=\left(\frac{\partial r_{\text{t}}\left( \mathbf{p}_{i}\right)}{\partial \mathbf{p}_{i}}-\frac{\left(\mathbf{p}_{i}-\mathbf{m}\left( \mathbf{p}_i\right)\right)^{\text{T}}}{\left \Vert \mathbf{p}_{i}-\mathbf{m}\left( \mathbf{p}_i\right) \right \Vert}\left( \mathbf{I}_{\text{2}}-\frac{\partial \mathbf{m}\left( \mathbf{p}_i\right)}{\partial  \mathbf{p}_i}\right) \right)
	\mathbf{v}_{\text{c},i}.
\end{equation*}
For ensuring $\mathcal{S}_{i}\cap \partial \mathcal{T}_{\mathcal{V}}=\varnothing$, at least it is required
$d{_{\text{t,}i}} >r_{\text{s}}$. However, this
constraint is not enough because the real distance from $\mathbf{p}_i$ to $\partial \mathcal{T}_{\mathcal{V}}$ is usually
smaller than $d{_{\text{t,}i}}$ as shown in Figure \ref{Threeaera2}. Hence a modified safety radius $r_{\text{s}}^{\prime }$ is proposed satisfying
\begin{equation}
	r_{\text{s}}^{\prime }=\inf_{\mathbf{x}\in \mathcal{V}}\text{dist}\left( \mathcal{C}%
	\left( \mathbf{x}\right) \cap \mathcal{S}_{\mathcal{T}_{\mathcal{V}%
	}},\partial \mathcal{T}_{\mathcal{V}}\right)  \label{rvs1}
\end{equation}%
and%
\begin{equation*}
	\mathcal{S}_{\mathcal{T}_{\mathcal{V}}}=\left \{  \mathbf{x}\in 
	\mathcal{T}_{\mathcal{V}}:\text{dist}\left( \mathbf{x},\partial 
	\mathcal{T}_{\mathcal{V}}\right) \geq r_{\text{s}}\right \}.
\end{equation*}
Then there exists the following proposition.

\textbf{Proposition 2}. For any $\mathbf{p}_{i}\in \mathcal{
	T}_{\mathcal{V}},$ if there exists
\begin{equation}
	r_{\text{t}}\left( \mathbf{p}_{i}\right) - \left \Vert \mathbf{p}_{i}-\mathbf{m}\left( \mathbf{p}_i\right) \right \Vert >r_{\text{s}}^{\prime},  \label{rvs}
\end{equation}
then it is obtained that $\mathcal{S}_{i}\subset \mathcal{\mathcal{T}_{\mathcal{V}}}$. 

\textit{Proof}. If (\ref{rvs}) holds, then $\mathbf{p}_{i}\in \mathcal{C}\left( \mathbf{p}_{i}\right) \cap \mathcal{S}_{\mathcal{T}_{\mathcal{V}}},$ namely $%
\mathbf{p}_{i}\in \mathcal{S}_{\mathcal{T}_{\mathcal{V}}}.$ This
implies that dist$\left( \mathbf{p}_{i},\partial \mathcal{T}_{\mathcal{V}%
}\right) \geq r_{\text{s}}$, namely $\mathcal{S}_{i}\subset \mathcal{%
	\mathcal{T}_{\mathcal{V}}}.$ $\square $

Then, a unified barrier function for the $i$th robot keeping within the tube is defined as
\begin{equation}
	V_{\text{t},i}=\frac{k_{3}\sigma _{\text{t}}\left(d{_{\text{t,}i}}\right) }{\left( 1+\epsilon _{\text{t}}\right) d{_{\text{t,}i}}
		-r_{\text{s}}^\prime s\left( \frac{d{_{\text{t,}i}} }{r_{\text{s}}^\prime},\epsilon _{\text{s}}\right) }, \label{Vti}
\end{equation}
where $\epsilon _{\text{t}}>0$. Here, the smooth function $\sigma \left( \cdot \right) $ in (\ref{zerofunction}) is defined as $\sigma _{\text{t}}\left( x\right) \triangleq \sigma \left( x,r_{\text{s}}^{\prime },r_{\text{a}}\right) $.
The function $V_{\text{t},i}$ has similar properties to $V_{\text{m},ij}$.
\begin{enumerate}
	\item[(i)] $\partial V_{\text{t},i} / \partial d{_{\text{t,}i}}\leq 0$ as $V_{\text{t},i}$ is a nonincreasing function with respect to $d{_{\text{t,}i}}$.
	\item[(ii)] If $d{_{\text{t,}i}}>r_\text{a}$, namely $\mathcal{A}_{i}\cap 	\partial \mathcal{T}_{\mathcal{V}}\neq \varnothing$, then $V_{\text{t},i}=0$ and $\partial V_{\text{t},i} / \partial d{_{\text{t,}i}}= 0$; if $V_{\text{t},i}=0$, then $d{_{\text{t,}i}}>r_\text{a}$.
	\item[(iii)] If $d{_{\text{t,}i}}<r_\text{s}^{\prime}$, namely $\mathcal{S}_{i}\cap 	\partial \mathcal{T}_{\mathcal{V}}\neq \varnothing$, then there exists a sufficiently small $\epsilon_{\text{s}}>0$ such that 
	\begin{equation*}
		V_{\text{t},ij}=\frac{k_2}{\epsilon_{\text{t}}d{_{\text{t,}i}}}\geq \frac{k_2}{\epsilon_{\text{t}}r_\text{s}^{\prime}}.
	\end{equation*}
\end{enumerate}
The objective of the designed velocity command is to make $V_{\text{t},i}$ zero or as small as possible. This implies that $d{_{\text{t,}i}} >r_{\text{s}}^{\prime }$, namely the $i$th robot will keep within the tube. 

\subsubsection{Modified Swarm Controller with a Non-Potential Term}
The modified swarm controller is proposed as 
\begin{equation}
	\mathbf{v}_{\text{c},i}=\mathbf{v}_\text{mdf}\left( \mathcal{T}_{\mathcal{V}},\mathbf{p}_{i},
	\mathbf{p},r_\text{s}^{\prime }\right), \label{modifiedcontroller}
\end{equation}
where
\begin{align*}
	&\mathbf{v}_{\text{mdf}}\left( \mathcal{T}_{\mathcal{V}},\mathbf{p}_{i},
	\mathbf{p},r_\text{s}^{\prime }\right) \triangleq -\text{sat}\left( \underset{\text{Line
			Approaching}}{\underbrace{-v_{\text{m},i}\mathbf{t}_{\text{c}}\left( \mathbf{p}_{i}\right) }}\right. \notag \\
	&\left.+{\underset{\text{Robot Avoidance}}{\underbrace{%
				\underset{j\in \mathcal{N}_{\text{m},i}}{\overset{}{\sum }}-b_{ij}%
				\tilde{\mathbf{p}}_{\text{m,}ij}}}}+\underset{\text{Virtual Tube Keeping}}{
		\underbrace{	\left(\mathbf{I}_{2}-\mathbf{t}_{\text{c}}\left(  \mathbf{p}{_{i}}\right)\mathbf{t}_{\text{c}}^{\text{T}}\left(  \mathbf{p}{_{i}}\right) \right)\mathbf{c}_{i}}},{v_{\text{m},i}}\right) .
\end{align*}
Here $\left(\mathbf{I}_2-\mathbf{t}_{\text{c}}\left( \mathbf{p}_{i}\right)\mathbf{t}^{\text{T}}_{\text{c}}\left( \mathbf{p}_{i}\right)\right)\mathbf{c}_i$ is the modified tube keeping term and $\mathbf{c}_i$ is expressed as
\begin{equation*}
	\mathbf{c}_{i}=\frac{\partial V_{\text{t},i}}{\partial d{_{\text{t,}i}}}\left(\frac{\partial r_{\text{t}}\left( \mathbf{p}_{i}\right)}{\partial \mathbf{p}_{i}}-\frac{\left(\mathbf{p}_{i}-\mathbf{m}\left( \mathbf{p}_i\right)\right)^{\text{T}}}{\left \Vert \mathbf{p}_{i}-\mathbf{m}\left( \mathbf{p}_i\right) \right \Vert}\left( \mathbf{I}_{\text{2}}-\frac{\partial \mathbf{m}\left( \mathbf{p}_i\right)}{\partial  \mathbf{p}_i}\right) \right)^{\text{T}}. 
\end{equation*}
The non-potential term\footnote{The non-potential term refers to a non-conservative vector field with no corresponding scalar potential function. In vector calculus, a non-conservative vector field is a vector field that is not the gradient of any scalar function \cite{marsden2003vector}. A non-conservative vector field is also rotational with its curl non-zero. } $\left(\mathbf{I}_2-\mathbf{t}_{\text{c}}\left( \mathbf{p}_{i}\right)\mathbf{t}^{\text{T}}_{\text{c}}\left( \mathbf{p}_{i}\right)\right)\mathbf{c}_i$ is used to imitate the performance of  $\partial V_{\text{tl},i}/\partial \mathbf{p}_i$ and $\partial V_{\text{tr},i}/\partial \mathbf{p}_i$.

\begin{figure}
	\centering
	\includegraphics[width=\columnwidth]{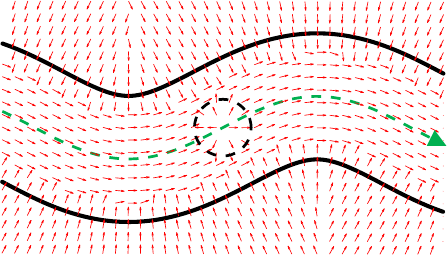}
	\caption{Vector field of a curve virtual tube with the modified controller \eqref{modifiedcontroller}.}
	\label{MVF}
\end{figure}

Consider a scenario that a robot is moving within a curve virtual tube, in the middle of which there exists another robot. Figure \ref{MVF} shows the vector field of this curve virtual tube with the modified swarm controller \eqref{modifiedcontroller}.

\textbf{Remark 5}. The term $\mathbf{c}_{i}$ is the gradient of $V_{\text{t},i}$, namely $\mathbf{c}_{i}=\left(\partial V_{\text{t},i}/\partial \mathbf{p}_i\right)^{\text{T}}$, which is always orthogonal to one of the boundary curve of $\partial \mathcal{T}_{\mathcal{V}}$. And $\left(\mathbf{I}_2-\mathbf{t}_{\text{c}}\left( \mathbf{p}_{i}\right)\mathbf{t}^{\text{T}}_{\text{c}}\left( \mathbf{p}_{i}\right)\right)\mathbf{c}_i$ is a non-potential velocity command component, which is always orthogonal to $\mathbf{t}_{\text{c}}\left( \mathbf{p}_{i}\right)$. For avoiding deadlock, directly applying $\mathbf{c}_{i}$ in (\ref{control_highway_dis}) is not feasible. Hence the use of the single panel method is necessary. For all $\mathbf{p} \in  \mathcal{V},l\left(\mathbf{p}\right)>l\left(\mathbf{p}^{-}_\text{a}\left(\mathbf{p}_i\right)\right),l\left(\mathbf{p}\right)<l\left(\mathbf{p}^{+}_\text{a}\left(\mathbf{p}_i\right)\right)$, if 
$\left \vert l_\text{r}\left(\mathbf{p}\right)-l_\text{l}\left(\mathbf{p}\right)\right \vert$ and $\left \vert r_\text{r}\left(\mathbf{p}\right)-r_\text{l}\left(\mathbf{p}\right)\right \vert$ are both very large, then the orientation changes of $\partial V_{\text{tl},i}/\partial \mathbf{p}_i$ and $\partial V_{\text{tr},i}/\partial \mathbf{p}_i$ inside the tube may be negligibly small. We can choose appropriate values of 
$l_\text{r}\left(\mathbf{p}\right),l_\text{l}\left(\mathbf{p}\right),r_\text{r}\left(\mathbf{p}\right),r_\text{l}\left(\mathbf{p}\right)$, so that 
the orientations of $\partial\phi_\text{l}\left(\mathbf{p}_i,\mathbf{p}\right)/\partial \mathbf{p}_i$ and $\partial\phi_\text{r}\left(\mathbf{p}_i,\mathbf{p}\right)/\partial \mathbf{p}_i$ are both approximately orthogonal to $\mathbf{t}_{\text{c}}\left( \mathbf{p}_{i}\right)$. Hence $\partial V_{\text{tl},i}/\partial \mathbf{p}_i$ and $\partial V_{\text{tr},i}/\partial \mathbf{p}_i$ are also both approximately orthogonal to $\mathbf{t}_{\text{c}}\left( \mathbf{p}_{i}\right)$, which explains why $\left(\mathbf{I}_2-\mathbf{t}_{\text{c}}\left( \mathbf{p}_{i}\right)\mathbf{t}^{\text{T}}_{\text{c}}\left( \mathbf{p}_{i}\right)\right)\mathbf{c}_i$ can imitate $\partial V_{\text{tl},i}/\partial \mathbf{p}_i$ and $\partial V_{\text{tr},i}/\partial \mathbf{p}_i$.

\textbf{Remark 6}. Compared with $V_{\text{tl},i}$, $V_{\text{tr},i}$ in (\ref{Vtli}), (\ref{Vtri}), the unified barrier function $V_{\text{t},i}$ in (\ref{Vti}) has its unique advantage of the broader application. In practice, the case such as $\text{dist}\left(\mathbf{p}_i,\partial \mathcal{T}_\mathcal{V}\right)<r_\text{s}$
may still happen in practice due to unpredictable uncertainties violating the assumptions. Under this circumstance, the potential functions $V_{\text{tl},i}$, $V_{\text{tr},i}$
have computation errors, while $V_{\text{t},i}$ still works well and the modified tube keeping term $\left(\mathbf{I}_2-\mathbf{t}_{\text{c}}\left( \mathbf{p}_{i}\right)\mathbf{t}^{\text{T}}_{\text{c}}\left( \mathbf{p}_{i}\right)\right)\mathbf{c}_i$ dominates the velocity command $\mathbf{v}_{\text{mdf}}\left( \mathcal{T}_{\mathcal{V}},\mathbf{p}_{i},
\mathbf{p},r_\text{s}^{\prime }\right) $ in \eqref{modifiedcontroller}, which implies that $	d{_{\text{t,}i}}$ will be increased very fast so that the $i$th robot can keep away from the tube boundary immediately.

\section{Simulation and Experiment}

In this section, simulations and experiments are given to show the effectiveness of the proposed method. A video about simulations and experiments is available on \href{https://youtu.be/NcjsCOZoEIw}{https://youtu.be/NcjsCOZoEIw} and \href{http://rfly.buaa.edu.cn}{http://rfly.buaa.edu.cn}.

\subsection{Simulation with the Single Integral Model at Different Maximum Speeds}

In this part, the validity and feasibility of the proposed method is numerically verified in a simulation. Consider a scenario that one robotic swarm composed of $M=20$ robots passes through a predefined curve virtual tube. All robots satisfy the model in (\ref{SingleIntegral}). The swarm controller in (\ref{modifiedcontroller}) is applied to guide these robots.

\begin{figure}
	\centering
	\includegraphics[width=\columnwidth]{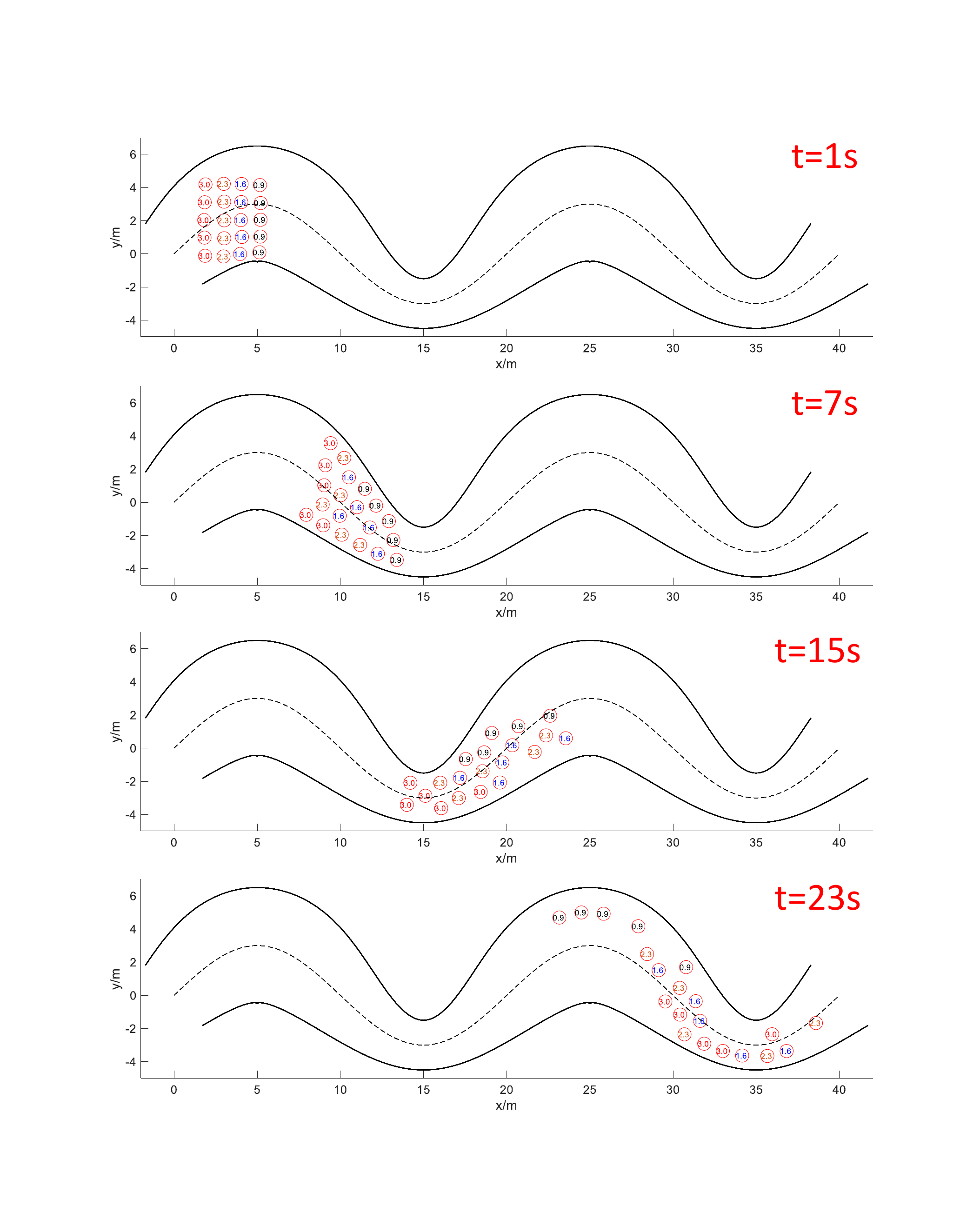}
	\caption{Single integral model simulation snapshots.}
	\label{simulation_diff1}
\end{figure}
\begin{figure}
	\centering
	\includegraphics[width=\columnwidth]{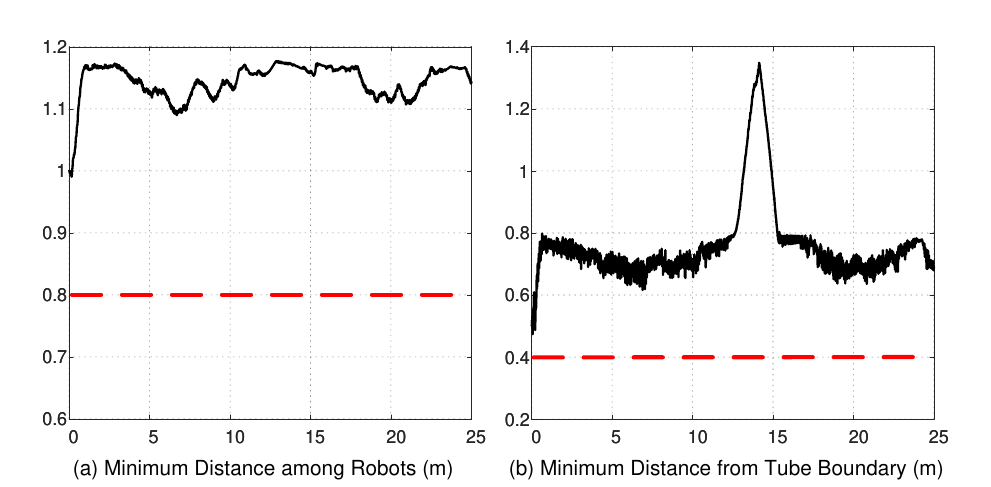}
	\caption{Minimum distance among robots and minimum distance from the tube boundary of all robots in the single integral model simulation.}
	\label{simulation_diff2}
\end{figure}

The parameters and initial conditions of the simulation are set as follows. The generating curve is designed manually as a sinusoidal curve in the middle of the curve virtual tube. The width of the tube also changes along the generating curve. The control parameters are $k_2=k_3=1$, $\epsilon_{\text{m}}=\epsilon_{\text{t}}=\epsilon_{\text{s}}=10^{-6}$. All robots with the safety radius $r_\text{s} = 0.4\text{m}$, the avoidance radius $r_\text{a}= 0.8\text{m}$ are arranged symmetrically in a rectangular space in the beginning. As shown in Figure \ref{simulation_diff1}, the boundaries of the safety area are represented by red circles. To show the ability to control different types of robots at the same time with our proposed method, the robot's maximum speed is set to four different constants:

\begin{equation*}
	v_{\text{m},i}=\left\{
	\begin{aligned}
		3.0\text{m/s}&  \quad i=1,\cdots,5& \\
		2.3\text{m/s}&  \quad i=6,\cdots,10& \\
		1.6\text{m/s}&  \quad i=11,\cdots,15& \\
		0.9\text{m/s}&  \quad i=16,\cdots,20& 
	\end{aligned}
	\right..
\end{equation*}
The corresponding maximum speed for each robot is shown with different colors in the center of the safety area. 

The simulation lasts 25 seconds and three snapshots are shown in Figure \ref{simulation_diff1}. It can be observed that the robots at the largest maximum speed $v_{\text{m},i}=3.0\text{m/s}$ are in the last column in the beginning. Then they have the trend to overtake other robots ahead. During the whole process, robots can change their relative positions freely instead of maintaining a fixed geometry structure. It is clear from Figure \ref{simulation_diff2}(a) that the minimum distance between any two robots
is always larger than $2r_\text{s}=0.8\text{m}$, which implies that there is no collision among robots. In Figure \ref{simulation_diff2}(b), the minimum distance from the tube boundary among all robots
keeps larger than $r_\text{s}=0.4\text{m}$ all the time. Therefore, all robots can avoid collision with each other and keep moving within the curve virtual tube under the proposed swarm controller. 


\subsection{Gazebo-based Simulation with Multicopter Flight Control Rigid Model}
\begin{figure}
	\centering
	\includegraphics[width=\columnwidth]{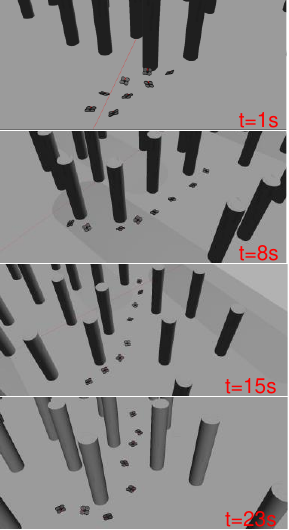}
	\caption{Gazebo-based simulation snapshot.}
	\label{Gazebo1}
\end{figure}
\begin{figure}
	\centering
	\includegraphics[width=\columnwidth]{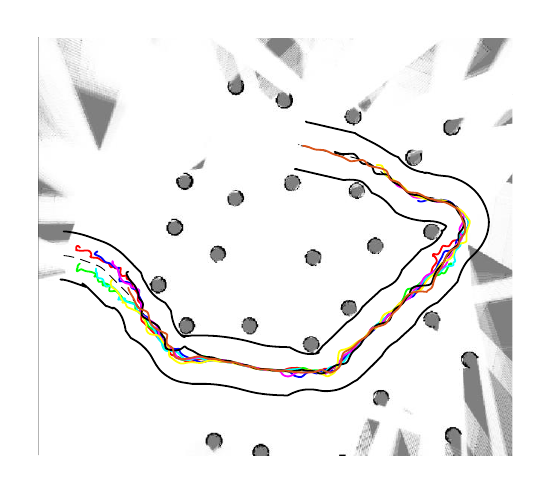}
	\caption{Trajectories of the swarm in the Gazebo-based simulation.}
	\label{GazeboOcto}
\end{figure}
\begin{figure}
	\centering
	\includegraphics[width=\columnwidth]{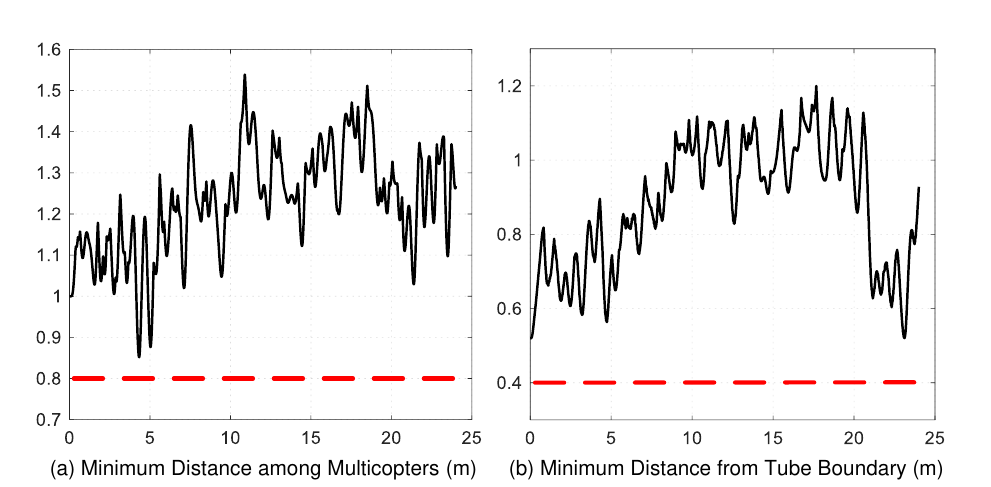}
	\caption{Minimum distance among multicopters and minimum distance from the tube boundary of all robots in the Gazebo-based simulation.}
	\label{Gazebocurve}
\end{figure}
In this part, the control performance of the proposed method is specially verified in a multicopter swarm. Different from previous simulations, a multicopter flight control rigid model \cite[pp. 126-127]{Quan(2017)} is applied to replace the single integral model in (\ref{SingleIntegral}). In other words, all multicopters are simulated by a six degree of freedom (DOF) nonlinear dynamic model. They are viewed as rigid bodies instead of mass points. A more precise model needs a more detailed and realistic simulation environment. A Gazebo-based simulation
environment modified from \cite{tordesillas2019faster} is used, which can simulate multiple multicopters at the same time. 

As described in (\ref{modifiedcontroller}), the velocity commands for all multicopters are designed by the proposed modified swarm controller. However, the control inputs for the multicopter flight control rigid model are the thrust and moments generated by four propellers. Hence, a inner-loop controller is designed for the multicopter to track the desired velocity. The control parameters are $k_2=k_3=1$, $\epsilon_{\text{m}}=\epsilon_{\text{t}}=\epsilon_{\text{s}}=10^{-6}$. All robots have the safety radius $r_\text{s} = 0.4\text{m}$, the avoidance radius $r_\text{a}= 0.8\text{m}$ and the maximum speed $v_{\text{m},i} = 3\text{m/s}$.

The simulation is arranged as a ``teach-and-repeat'' process. To simplify the simulation, all multicopters have the same desired altitude. Given four waypoints $\mathbf{p}_\text{wp1}=[20\ 0]^\text{T}\text{m}$, $\mathbf{p}_\text{wp2}=[32\ -8]^\text{T}\text{m}$, $\mathbf{p}_\text{wp3}=[41\ 0]^\text{T}\text{m}$, $\mathbf{p}_\text{wp4}=[31\ 7]^\text{T}\text{m}$, a first multicopter equipped with a simulated Hokuyo lidar takes off at $\mathbf{p}_\text{1}(0)=[17\ 0]^\text{T}\text{m}$ and then passes all waypoints. The traditional Vector Field Histogram (VHF) method \cite{borenstein1991vector} is used to navigate this multicopter and avoid collisions with the obstacles in the environment. In this process, an occupancy map representing the surrounding environment is created from the lidar data. Besides, the flight trajectory is recorded and then utilized to create the curve virtual tube, whose boundary is generated by searching for the closest obstacle in the map. There we have completed the ``teach'' process.

The next is the ``repeat'' process. We apply $M=8$ multicopters composing a multicopter swarm to pass through the curve virtual tube. The initial positions of these eight multicopters are $\mathbf{p}_\text{2}(0)=[18\ 0.5]^\text{T}\text{m}$, $\mathbf{p}_\text{3}(0)=[18\ -0.5]^\text{T}\text{m}$, $\mathbf{p}_\text{4}(0)=[19\ 0.2]^\text{T}\text{m}$, $\mathbf{p}_\text{5}(0)=[19\ -0.8]^\text{T}\text{m}$, $\mathbf{p}_\text{6}(0)=[20\ -0.5]^\text{T}\text{m}$, $\mathbf{p}_\text{7}(0)=[20\ -1.5]^\text{T}\text{m}$, $\mathbf{p}_\text{8}(0)=[21\ -1]^\text{T}\text{m}$, $\mathbf{p}_\text{9}(0)=[21\ -2]^\text{T}\text{m}$. The modified swarm controller (\ref{modifiedcontroller}) is used to guide this swarm.
The simulation lasts 23 seconds and four snapshots are shown in Figure \ref{Gazebo1}. The environment map and flight trajectories of these eight multicopters are shown in Figure \ref{GazeboOcto}. It is obvious that all multicopters keep flying within the curve virtual tube in the simulation. Similarly to the analysis in the last subsection, as shown in Figure \ref{Gazebocurve}(a), the minimum distance between any two multicopters is always larger than $2r_\text{s}=0.8\text{m}$, which implies that there is no collision in the swarm. Besides, as shown in Figure \ref{Gazebocurve}(b), the minimum distance from the tube boundary among all multicopters keeps larger than $r_\text{s}=0.4\text{m}$ all the time. Therefore, in this Gazebo-based simulation, all robots successfully avoid colliding with each other and keep flying in the curve virtual tube under the proposed swarm controller.

\subsection{Comparison of the Calculation Speed With CBF Method}
\begin{figure}
	\centering
	\includegraphics[width=\columnwidth]{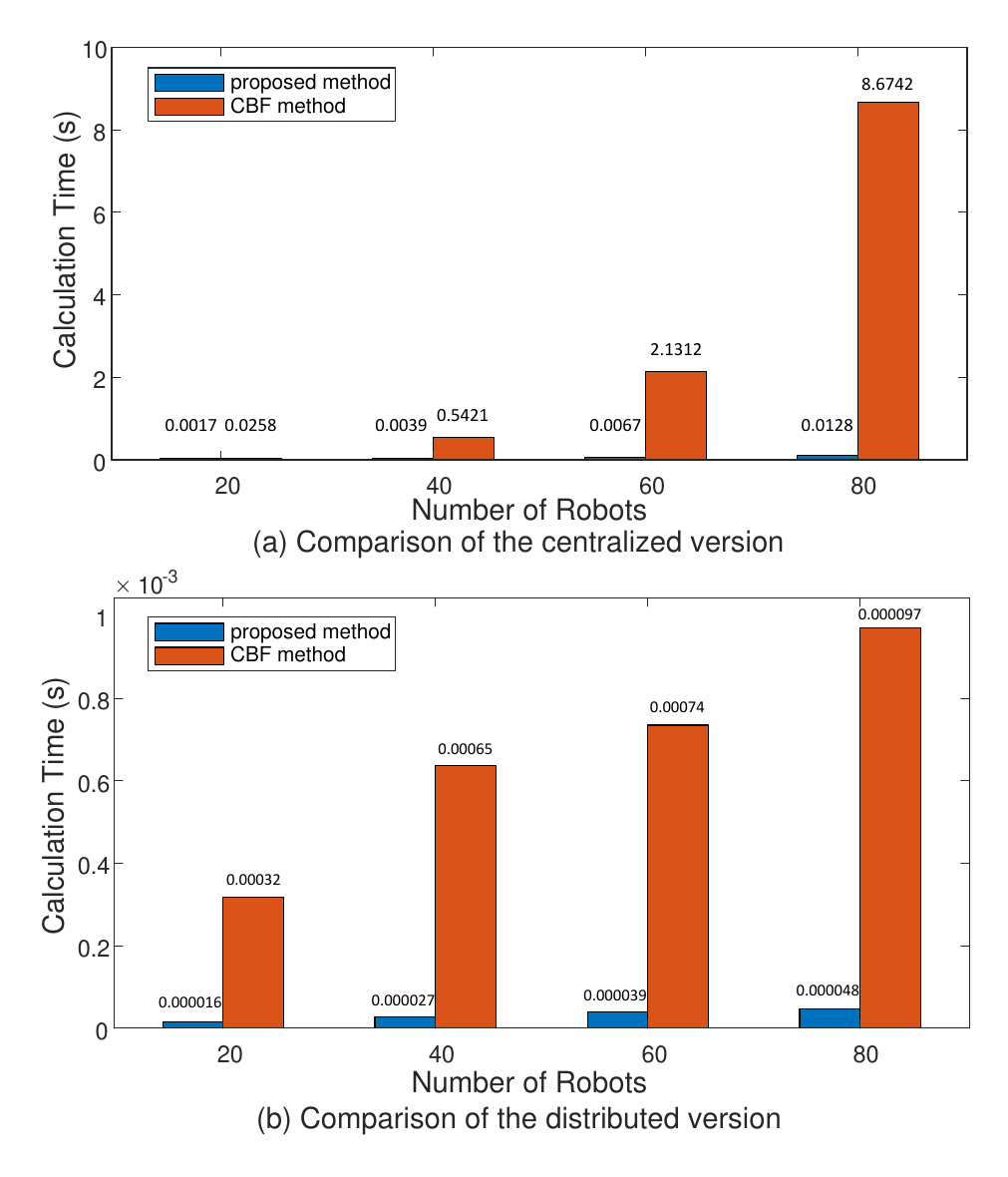}
	\caption{The calculation speed of our proposed method and CBF method.}
	\label{TimeCompare}
\end{figure}

In order to show the advantage of our proposed control method, we compare our method with the CBF method in the calculation speed. The controller (\ref{modifiedcontroller}) can be easily converted into a QP controller proposed in \cite{Wang(2017)}. The line approaching term corresponds to the nominal controller. The robot avoidance term and the tube keeping term corresponds to linear inequality constraints. As same as our method, the CBF method also has a distributed version. Here we compare the centralized and distributed versions of these two methods, respectively. The simulation is run on the same computer and record the average calculation time for both methods. Figure \ref{TimeCompare} shows that our method possesses a much higher calculation speed than the CBF method whether centralized or distributed. For the centralized version, the calculation time of the CBF method will increase rapidly when the number of robots increases, while the calculation time of our method only increases just a little. The reason is that for a dense and complex environment, inequality constraints of the CBF method may become contradictory, which leads to no solution to the QP problem. To deal with this problem, the relaxation variable is an option, but it brings the decline of safety. For the distributed version, the calculation time of both methods are limitedly affected by the increase of robots` number, which strongly expresses the advantage of the distributed methods.

\subsection{Real Experiment}
\begin{figure}
	\centering
	\includegraphics[width=\columnwidth]{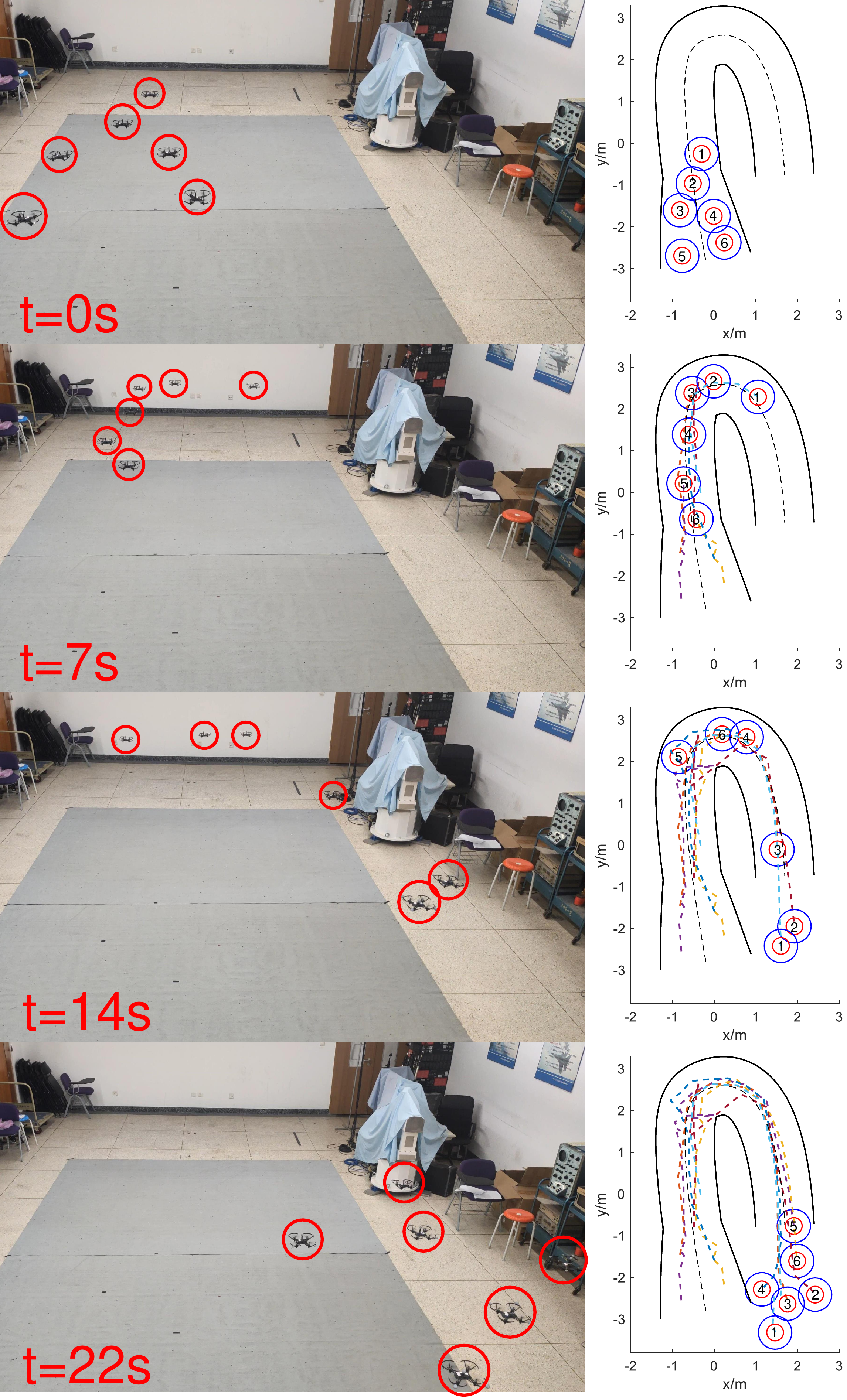}
	\caption{Experiment snapshot.}
	\label{experiment1}
\end{figure}

A real experiment is carried out in a laboratory room with $M=6$ Tello quadcopters and an OptiTrack motion capture system providing the positions and orientations of quadcopters. A laptop computer is connected to Tello quadcopters and OptiTrack with a local wireless network, running the proposed distributed controller (\ref{modifiedcontroller}). The generating curve is designed manually in the middle of the curve virtual tube, whose width also changes along the generating curve. The control parameters are $k_2=k_3=1$, $\epsilon_{\text{m}}=\epsilon_{\text{t}}=\epsilon_{\text{s}}=10^{-6}$.
All quadcopters have the safety radius $r_\text{s} = 0.2\text{m}$, the avoidance radius $r_\text{a} = 0.4\text{m}$ and the maximum speed $v_{\text{m},i}=0.5\text{m/s},i=1,\cdots,6$. As shown in Figure \ref{experiment1}, the boundaries of the safety and avoidance area are represented by red and blue circles respectively. 

\begin{figure}
	\centering
	\includegraphics[width=\columnwidth]{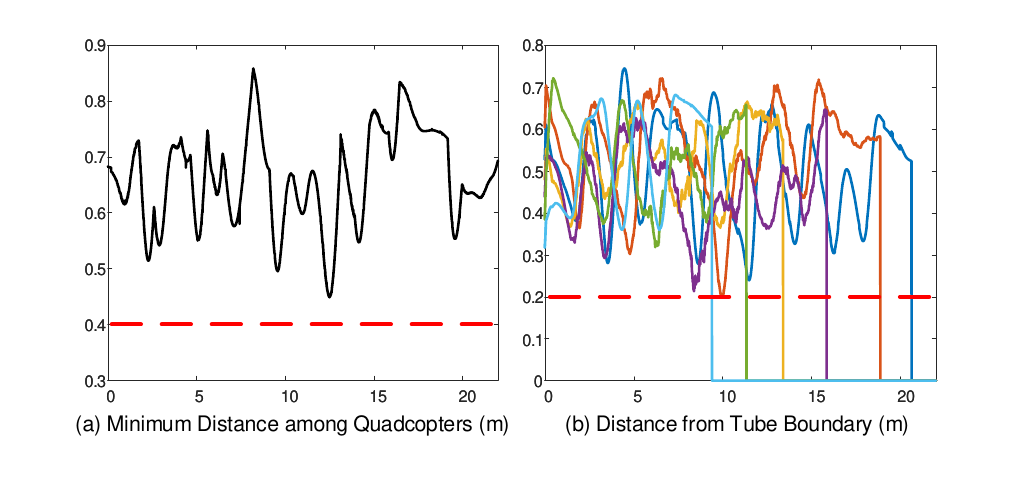}
	\caption{Minimum distance among quadcopters and distance from tube boundary. In plot (b), all curves approaching zero means that their corresponding quadcopters have moved across the finishing line of the curve virtual tube, which can be confirmed in the snapshot of $t=22\text{s}$ in Figure \ref{experiment1}.}
	\label{experiment3}
\end{figure}

The experiment lasts 22 seconds and four snapshots are shown in Figure \ref{experiment1}. As same as the numerical simulation, quadcopters can change their relative positions freely instead of maintaining a fixed geometry structure. It can be observed from Figure \ref{experiment3}(a) that the minimum distance between any two quadcopters is always larger than $2r_\text{s}=0.4\text{m}$, which implies that there is no collision among quadcopters. In Figure \ref{experiment3}(b), the distances from the tube boundary of all quadcopters keep larger than $r_\text{s}=0.2\text{m}$ when the quadcopters are inside the tube. 

%

\section{Conclusion}
The curve virtual tube passing through problem, which includes all robots passing through the tube, inter-robot collision avoidance and keeping within the tube, is proposed and then solved in this paper. Based on the artificial potential field method with a control saturation, practical distributed swarm control is proposed for multiple robots to pass through a curve virtual tube. Lyapunov-like
functions are designed elaborately, and formal analysis and
proofs are made to show that the curve virtual tube passing through problem can
be solved, namely all robots avoid collision with each other and keep within the tube in \emph{Lemma 1}, all robots pass through the tube without getting trapped in \emph{Theorem 1}. Simulations and experiments are given to show the effectiveness and performance of the proposed method under different kinds of conditions. To show the advantages of the proposed method over other algorithms
in terms of calculation speed of finding feasible solutions, the comparison between our method and CBF method is also presented.


\appendices
\section{Proof of Lemma 1}
Since we have 
\begin{equation*}
	\text{sat}\left( k_{1}l\left( \mathbf{x}\right) \eta \left( \mathbf{x}%
	\right) \mathbf{t}_{\text{c}}\left( \mathbf{x}\right) ,{v_{\text{m},i}}\right)
	={{\kappa }_{{v_{\text{m},i}}}}k_{1}l\left( \mathbf{x}\right) \eta \left( \mathbf{x}%
	\right) \mathbf{t}_{\text{c}}\left( \mathbf{x}\right)
\end{equation*}
where 
\begin{equation*}
	{{\kappa }_{{v_{\text{m},i}}}}=\left \{ 
	\begin{array}{c}
		1 \\ 
		\frac{{v_{\text{m},i}}}{k_{1}\eta \left( \mathbf{x}
			\right) \left \vert l\left( \mathbf{x}\right) \right \vert }
	\end{array}%
	\begin{array}{c}
		\left \vert k_{1}l\left( \mathbf{x}\right) \eta \left( \mathbf{x}
		\right) \right \vert \leq {v_{\text{m},i}} \\ 
		\left \vert k_{1}l\left( \mathbf{x}\right) \eta \left( \mathbf{x}
		\right) \right \vert >{v_{\text{m},i}}
	\end{array}
	\right. .
\end{equation*}
Then, the function (\ref{Vli0}) is rewritten as 
\begin{equation}
	V_{\text{li}}\left( \mathbf{y}\right) =\int_{\mathcal{V}_{\mathbf{y}}}{\kappa }_{v_{\text{m},i}}k_{1}l\left( \mathbf{x}\right) \eta \left( \mathbf{x}%
	\right) \mathbf{t}_{\text{c}}\left( \mathbf{x}\right)^{\text{T}}\text{d}\mathbf{x}.  \label{Vli10}
\end{equation}
With (\ref{dl}), the function (\ref{Vli10}) becomes 
\begin{align}
	V_{\text{li}}\left( \mathbf{y}\right) =&\int_{\mathcal{V}_{\mathbf{y}}}{{
			\kappa }_{{v_{\text{m},i}}}}k_{1}l\left( \mathbf{x}\right) \text{d}l\left( \mathbf{x}\right)   \notag\\
	=&\int_{\mathcal{V}_{\mathbf{y}}}\frac{{{\kappa }_{{v_{\text{m},i}}}}}{2k_{1}}%
	\text{d}\left \vert k_{1}l\left( \mathbf{x}\right) \right \vert ^{2}  \notag \\
	=&\int_{0}^{\left \vert k_{1}l\left( \mathbf{y}\right) \right \vert }\frac{{{\kappa }_{{%
					v_{\text{m},i}}}}}{2k_{1}}\text{d}z^{2}  \label{Vli11}
\end{align}%
where $z=\left \vert k_{1}l\left( \mathbf{x}\right) \right \vert .$ According to (\ref%
{Vli11}), since $\frac{{{\kappa }_{{v_{\text{m},i}}}}}{2k_{1}}>0,$ we have $V_{%
	\text{li}}\left( \mathbf{y}\right) >0$ if $\left \vert l\left( \mathbf{y}%
\right) \right \vert \neq 0$, and $V_{\text{li}}\left( \mathbf{y}\right) =0$
if and only if $\left \vert l\left( \mathbf{y}\right) \right \vert =0.$ 
When there exists $\left \vert k_{1}l\left( \mathbf{x}\right) \eta \left( \mathbf{x}%
\right) \right \vert >{v_{\text{m},i}}$, it is obtained that $\frac{{v_{\text{m},i}}}{k_{1}\eta \left( \mathbf{x}\right) \left \vert l\left( \mathbf{x}\right) \right \vert }<1$. Hence we have $\kappa _{v_{\text{m},i}}\leq1$ and
\begin{align}
	V_{\text{li}}\left( \mathbf{y}\right) \geq &\int_{0}^{\left \vert k_{1}l\left( \mathbf{y}\right)\right \vert }\frac{1}{2k_{1}}\frac{{v_{\text{m},i}}}{k_{1}\eta \left( \mathbf{x}\right) \left \vert l\left( \mathbf{x}\right) \right \vert }\text{d}z^{2}  \notag \\
	\geq &\int_{0}^{\left \vert k_{1}l\left( \mathbf{y}\right)\right \vert }\frac{{v_{%
				\text{m},i}}}{2k_{1}\eta _{\max }}\frac{1}{z}\text{d}z^{2}  \notag \\
	=&\frac{{v_{\text{m},i}}}{\eta _{\max }}\left \vert l\left( \mathbf{y}\right)\right
	\vert .  \label{Vli3}
\end{align}%
If $\left \Vert \mathbf{y}\right \Vert \rightarrow \infty ,$ then $\left
\Vert \mathbf{y-p}_{\text{f}}\right \Vert \rightarrow \infty .$ Since $\left
\Vert \mathbf{y-p}_{\text{f}}\right \Vert $ is the shortest distance from $%
\mathbf{y}$ to $\mathbf{p}_{\text{f}}$, we have $\left \vert l\left( \mathbf{y}\right)\right \vert \geq \left \Vert \mathbf{y-p}_{\text{f}}\right \Vert .$
Therefore, from (\ref{Vli3}), we have $V_{\text{li}}\left( \mathbf{y}\right)
\rightarrow \infty .$ This also implies that if $V_{\text{li}}\left( \mathbf{y}%
\right) $ is bounded, then $\left \Vert \mathbf{y}\right \Vert $ is bounded. 
$\square $
\section{Proof of Lemma 2}
The reason why these robots are able to avoid conflict with each other will be proved by contradiction. Without loss of generality, assume
that  $\left \Vert \tilde{\mathbf{p}}_{\text{m},ij_1}\left( t_{2}\right) \right \Vert =2r_{\text{s}}$ occurs at $t_2>0$ first, namely there is a conflict between the $i$th robot and the $j_{1}$th robot. Besides, we have $\left \Vert \tilde{\mathbf{p}}_{\text{m},ij}\left( t_{2}\right) \right \Vert >2r_{\text{s}}$ for $j\neq j_{1}$.
Consequently, $V_{\text{m},ij}{\left( t_2\right) \geq 0}$ if $j\neq
j_{1}.$ Since ${V}\left( 0\right) >0$ and ${{\dot{V}}\left( t\right) }\leq 0$%
, the function ${V}$ satisfies ${V}\left( t_2\right) \leq {V}\left(
0\right) ,$ $t\in \left[ 0,\infty \right) $. By the definition of ${{V},}$
we have $V{_{\text{m,}ij_{1}}}\left( t_2\right) \leq {V}\left( 0\right) .$
Given any $\epsilon _\text{rs}>0,$ there exists a $%
\epsilon _{\text{s}}>0,$ such that $s\left( 1,\epsilon _{\text{s}}\right) =1-\epsilon _\text{rs}$.
Then, at the time $t_2,$ the denominator of $V_{\text{m},ij_{1}}$
defined in (\ref{Vmij}) is 
\begin{align*}
	& \left( 1+\epsilon _{\text{m}}\right) \left \Vert \tilde{\mathbf{p}}_{\text{m},ij_1}\left( t_{2}\right) \right \Vert -2r_{\text{s}%
	}s\left( \frac{\left \Vert \tilde{\mathbf{p}}_{\text{m},ij_1}\left( t_{2}\right) \right \Vert}{2r_{\text{s}}},\epsilon _{\text{s}%
	}\right)  \notag \\
	& =2r_{\text{s}}\left( 1+\epsilon _{\text{m}}\right) -2r_{\text{s}}\left(
	1-\epsilon _\text{rs}\right)  \notag \\
	& =2r_{\text{s}}\left( \epsilon _{\text{m}}+\epsilon _\text{rs}\right)
	\label{bound1}
\end{align*}%
where $\epsilon _\text{rs}>0$ can be sufficiently small if $\epsilon _{\text{s}}$
is sufficiently small. According to the definition
in (\ref{Vmij}), we have 
\begin{equation*}
	\frac{1}{2r_{\text{s}}\left( \epsilon _{\text{m}}+\epsilon _\text{rs}\right) }=%
	\frac{V{_{\text{m,}ij_{1}}}\left( t_2\right) }{k_{2}}\leq \frac{{V}%
		\left( 0\right) }{k_{2}}  \label{fact}
\end{equation*}%
where $\sigma _{_{\text{m}}}\left( \left \Vert \tilde{\mathbf{p}}_{\text{m},ij_1} \right \Vert \right) =1$ is used. Consequently, ${{V}\left(
	0\right) }$ is \emph{unbounded} as $\epsilon _{\text{m}}\rightarrow 0\ $and $%
\epsilon _\text{rs}\rightarrow 0.$ On the other hand, for any $j$, we have $\left \Vert \tilde{\mathbf{p}}_{\text{m},ij}\left(0\right) \right \Vert>2r_{\text{s}}$ by \textit{Assumption 3}. Let $\left \Vert \tilde{\mathbf{p}}_{\text{m},ij}\left(0\right) \right \Vert=2r_{%
	\text{s}}+{\varepsilon _{\text{m,}ij},}$ ${\varepsilon _{\text{m,}ij}}>0$.
Then, at the time $t=0,$ the denominator of $V_{\text{m},ij}$ defined in (\ref%
{Vmij}) is 
\begin{align*}
	& \left( 1+\epsilon _{\text{m}}\right) \left \Vert \tilde{\mathbf{p}}_{\text{m},ij}\left(0\right) \right \Vert -2r_{\text{s}}s\left( \frac{\left \Vert \tilde{\mathbf{p}}_{\text{m},ij}\left(0\right) \right \Vert}{2r_{\text{s}}},\epsilon _{\text{s}}\right) \\
	& \geq \left( 1+\epsilon _{\text{m}}\right) \left( 2r_{\text{s}}+{%
		\varepsilon _{\text{m,}ij}}\right) -2r_{\text{s}}\bar{s}\left( \frac{\left \Vert \tilde{\mathbf{p}}_{\text{m},ij}\left(0\right) \right \Vert}{%
		2r_{\text{s}}}\right) \\
	& =2r_{\text{s}}\epsilon _{\text{m}}+\left( 1+\epsilon _{\text{m}}\right) {%
		\varepsilon _{\text{m,}ij}}.
\end{align*}%
Then we have
\begin{equation*}
	V{_{\text{m,}ij}}\left( 0\right) \leq \frac{k_{2}}{2r_{\text{s}}\epsilon _{%
			\text{m}}+\left( 1+\epsilon _{\text{m}}\right) {\varepsilon _{\text{m,}ij}}}.
\end{equation*}%
Consequently, $V{_{\text{m,}ij}}\left( 0\right) $ is still bounded as $%
\epsilon _{\text{m}}\rightarrow 0\ $no matter what $\epsilon _\text{rs}\ $is.
According to the definition of ${V}\left( 0\right) ,$ ${V}\left( 0\right) $
is still \emph{bounded} as $\epsilon _{\text{m}}\rightarrow 0\ $and $%
\epsilon _\text{rs}\rightarrow 0.$ This is a contradiction. Thus there exists $\left \Vert \tilde{\mathbf{p}}_{\text{m},ij}\left(t\right) \right \Vert>2r_{\text{s}}$
for $i,j=1,\cdots ,N,i\neq j $, $t\in \left[ 0,\infty \right) .$ Therefore, the robot
can avoid any other robot by the velocity command (\ref{control_highway_dis}).

The reason why a robot can stay within the curve virtual tube is similar to the above
proof. It can be proved by contradiction as well. Without loss of
generality, assume that $\text{dist}\left(\mathbf{p}_i\left( t_3\right),\partial \mathcal{T}_{\mathcal{V}}\right)=r_\text{s}$ occurs at $t_3>0$, namely a conflict happens. Similar to the above
proof, we can also get a contradiction. $\square $

\section{Proof of Theorem 1} 

According to \textit{Lemma 2}, these robots are able to avoid conflict with each other and keep within the curve virtual tube, namely $\mathcal{S}_{i}\left(t\right)\cap \mathcal{S}_{j}\left(t\right)=\varnothing,$ $\mathcal{S}_{i}\left(t\right)\cap \partial \mathcal{T}_{\mathcal{V}}=\varnothing$, $t\in \lbrack 0,\infty )$ for all ${{\mathbf{p}}_{i}(0)}$, $i,j=1,\cdots ,M$ and $i \neq j$. In the following, the reason why the $i$th robot is able to approach the cross section $\mathcal{C}\left( {{\mathbf{p}}_{\text{f}}}\right) $ is given. As the function $V$ is not a Lyapunov function, the \textit{invariant set theorem} \cite[p. 69]{Slotine(1991)} is used to do the analysis here.

\begin{enumerate}
	\item[(i)] Firstly, we will study the property of the function $V$. Let $%
	\Omega=\left
	\{ \mathbf{p}_{1},\cdots, \mathbf{p}_{M}: {V}\leq l\right \} ,$ $l>0. $ According to \textit{Lemma 2}, there exists $V_{\text{m},ij},$ $V_{\text{tl},i}$, $V_{\text{tr},i}>0.$ Therefore, ${V}\leq l$ implies $\sum_{i=1}^{M}V_{\text{l},i}\leq l.$ Furthermore, according to \textit{%
		Lemma 1(iii)}, $\Omega$ is bounded. When $\left \Vert \left[\mathbf{p}
	_{1},\cdots, \mathbf{p}_{M}\right]\right \Vert \rightarrow \infty,$ then $\sum_{i=1}^{M}V_{\text{l},i}\rightarrow \infty$ according to \textit{Lemma 1(ii)},
	namely ${V}\rightarrow \infty$. Therefore the function $V$ satisfies the
	condition that the invariant set theorem requires.
	
	\item[(ii)] Secondly, we will find the largest invariant set. It is obtained that ${\dot{V}}={0}$
	if and only if 
	\begin{align*}
		&{{\kappa }_{v_{\text{m},i}}} k_{1}l\left( \mathbf{p}{_{i}}\right) \eta
		\left( \mathbf{p}{_{i}}\right) \mathbf{t}_{\text{c}}\left( 
		\mathbf{p}{_{i}}\right) -\underset{j=1,j\neq
			i}{\overset{M}{\sum}}b_{ij}\tilde{\mathbf{p}}_{\text{m},ij}\\
		&+\left(\frac{\partial V_{\text{tl},i}}{\partial \mathbf{p}_i }\right)^{\text{T}}+\left(\frac{\partial V_{\text{tr},i}}{\partial \mathbf{p}_i }\right)^{\text{T}}=\mathbf{0},
	\end{align*}
	where $i=1,\cdots,M$. Then we have $\mathbf{v}_{\text{c},i}=\mathbf{0\ }$according
	to ({\ref{control_highway_dis}}). Consequently, the system cannot get ``stuck'' at an equilibrium point other than $\mathbf{v}_{\text{c},i}=\mathbf{0}$. 
	
	\item[(iii)] Finally, we will prove that no robot will get ``stuck''. Let the $1$st robot be ahead of the robotic swarm, namely it is the closest to the finishing line $\mathcal{C}\left( {{\mathbf{p}}_{\text{f}}}\right) $. And the other robots are at back of $\mathcal{C}\left( {{\mathbf{p}}_{\text{1}}}\right)$. When there exists $\mathbf{v}_{\text{c},1}=\mathbf{0}$, we examine the following equation related to the 1st robot that 
	\begin{align}
		&{{\kappa }_{v_{\text{m},1}}}k_{1}l\left( \mathbf{p}{_{1}}\right) \eta \left( 
		\mathbf{p}{_{1}}\right) \mathbf{t}_{\text{c}}\left( \mathbf{p}{_{1}}\right) -
		\sum_{j=2}^{M}b_{1j}\mathbf{\tilde{p}}_{\text{m,}1j} \notag\\
		&+\left(\frac{\partial V_{\text{tl},1}}{\partial \mathbf{p}_1 }\right)^{\text{T}}+\left(\frac{\partial V_{\text{tr},1}}{\partial \mathbf{p}_1 }\right)^{\text{T}}=\mathbf{0}. \label{equilibriumTh5_1st}
	\end{align}
	Since the 1st robot is ahead, we have
	\begin{equation}
		\mathbf{t}_{\text{c}}^{\text{T}}\left( \mathbf{p}{_{1}}\right) \mathbf{%
			\tilde{p}}_{\text{m,}1j}\geq 0  \label{1st}
	\end{equation}%
	where ``$=$'' \ holds if and only if the $j$th robot is as ahead as the $1$st one. Then, multiplying the term $\mathbf{t}_{\text{c}}^{\text{T}}\left(  \mathbf{p}{_{1}}\right)$ at the left side of (\ref{equilibriumTh5_1st}) results in
	\begin{align*}
		&{{\kappa }_{v_{\text{m},1}}}k_{1}l\left( \mathbf{p}{_{1}}\right) \eta \left( 
		\mathbf{p}{_{1}}\right) \\
		=&\mathbf{t}_{\text{c}}^{\text{T}}\left( \mathbf{p}{_{1}}\right)\sum_{j=2}^{M}b_{1j}\mathbf{\tilde{p}}_{\text{m,}1j}-\mathbf{t}_{\text{c}}^{\text{T}}\left(\mathbf{p}_1\right)\left(\frac{\partial V_{\text{tl},1}}{\partial \mathbf{p}_1 }\right)^{\text{T}}\\
		&-\mathbf{t}_{\text{c}}^{\text{T}}\left(\mathbf{p}_1\right)\left(\frac{\partial V_{\text{tr},1}}{\partial \mathbf{p}_1 }\right)^{\text{T}}\geq 0
	\end{align*}
	where  (\ref{DirL}), (\ref{DirR}), (\ref{1st}) are used. Then we have $l\left( \mathbf{p}{_{1}}\right)\geq0$.
	Since $l\left( \mathbf{p}{_{1}}\left( 0\right) \right) <0$ according to 
	\textit{Assumption 2}, owing to the continuity, given ${\epsilon }_{\text{0}%
	}>0,$ there must exist a time $t_{11}>0$ such that 
	\begin{equation*}
		l\left( \mathbf{p}{_{1}}\right) \geq -{\epsilon }_{\text{0}}
	\end{equation*}%
	when $t\geq t_{11}.$ At the time $t_{11},$ the $1$st robot is removed from the curve virtual tube according to \textit{Assumption 4}. The rest of problem is to consider the $M-1$ robots, namely $2$nd, $3$rd, ..., $M$th robots. We can repeat the analysis above to
	conclude this proof. $\square $
\end{enumerate}

\bibliographystyle{IEEEtran}
\bibliography{curvetunnel}

\begin{thebibliography}{10}
\providecommand{\url}[1]{#1}
\csname url@samestyle\endcsname
\providecommand{\newblock}{\relax}
\providecommand{\bibinfo}[2]{#2}
\providecommand{\BIBentrySTDinterwordspacing}{\spaceskip=0pt\relax}
\providecommand{\BIBentryALTinterwordstretchfactor}{4}
\providecommand{\BIBentryALTinterwordspacing}{\spaceskip=\fontdimen2\font plus
\BIBentryALTinterwordstretchfactor\fontdimen3\font minus
  \fontdimen4\font\relax}
\providecommand{\BIBforeignlanguage}[2]{{%
\expandafter\ifx\csname l@#1\endcsname\relax
\typeout{** WARNING: IEEEtran.bst: No hyphenation pattern has been}%
\typeout{** loaded for the language `#1'. Using the pattern for}%
\typeout{** the default language instead.}%
\else
\language=\csname l@#1\endcsname
\fi
#2}}
\providecommand{\BIBdecl}{\relax}
\BIBdecl

\bibitem{Chung(2018)}
S.~J. Chung, A.~A. Paranjape, P.~Dames, S.~Shen, and V.~Kumar, ``A survey on
  aerial swarm robotics,'' \emph{IEEE Transactions on Robotics}, vol.~34,
  no.~4, pp. 837--855, 2018.

\bibitem{ding2021epsilon}
W.~Ding, L.~Zhang, J.~Chen, and S.~Shen, ``Epsilon: An efficient planning
  system for automated vehicles in highly interactive environments,''
  \emph{IEEE Transactions on Robotics}, 2021.

\bibitem{ren2008distributed}
W.~Ren and N.~Sorensen, ``Distributed coordination architecture for multi-robot
  formation control,'' \emph{Robotics and Autonomous Systems}, vol.~56, no.~4,
  pp. 324--333, 2008.

\bibitem{Cheung(2019)}
Y.~Cheung, J.~H. Chung, and N.~P. Coleman, ``Semi-autonomous formation control
  of a single-master multi-slave teleoperation system,'' in \emph{2009 IEEE
  Symposium on Computational Intelligence in Control and Automation}.\hskip 1em
  plus 0.5em minus 0.4em\relax IEEE, 2009, pp. 117--124.

\bibitem{Oh(2015)}
K.~K. Oh, M.~C. Park, and H.~S. Ahn, ``A survey of multi-agent formation
  control,'' \emph{Automatica}, vol.~53, pp. 424--440, 2015.

\bibitem{Khan(2016)}
M.~U. Khan, S.~Li, Q.~Wang, and Z.~Shao, ``Distributed multirobot formation and
  tracking control in cluttered environments,'' \emph{ACM Transactions on
  Autonomous and Adaptive Systems (TAAS)}, vol.~11, no.~2, pp. 1--22, 2016.

\bibitem{Zhao(2019)}
S.~Zhao and D.~Zelazo, ``Bearing rigidity theory and its applications for
  control and estimation of network systems: Life beyond distance rigidity,''
  \emph{IEEE Control Systems Magazine}, vol.~39, no.~2, pp. 66--83, 2019.

\bibitem{Saska(2020)}
M.~Saska, D.~Hert, T.~Baca, V.~Kr{\'a}tk{\`y}, and T.~P. do~Nascimento,
  ``Formation control of unmanned micro aerial vehicles for straitened
  environments.'' \emph{Auton. Robots}, vol.~44, no.~6, pp. 991--1008, 2020.

\bibitem{dong2020controlling}
X.~Dong and M.~Sitti, ``Controlling two-dimensional collective formation and
  cooperative behavior of magnetic microrobot swarms,'' \emph{The International
  Journal of Robotics Research}, vol.~39, no.~5, pp. 617--638, 2020.

\bibitem{xu2020affine}
Y.~Xu, S.~Zhao, D.~Luo, and Y.~You, ``Affine formation maneuver control of
  high-order multi-agent systems over directed networks,'' \emph{Automatica},
  vol. 118, p. 109004, 2020.

\bibitem{Hao(2016)}
N.~D. Hao, B.~Mohamed, H.~Rafaralahy, and M.~Zasadzinski, ``Formation of
  leader-follower quadrotors in cluttered environment,'' in \emph{2016 American
  Control Conference (ACC)}.\hskip 1em plus 0.5em minus 0.4em\relax IEEE, 2016,
  pp. 6477--6482.

\bibitem{Mellinger(2012)}
D.~Mellinger, A.~Kushleyev, and V.~Kumar, ``Mixed-integer quadratic program
  trajectory generation for heterogeneous quadrotor teams,'' in \emph{2012 IEEE
  International Conference on Robotics and Automation}.\hskip 1em plus 0.5em
  minus 0.4em\relax IEEE, 2012, pp. 477--483.

\bibitem{Augugliaro(2012)}
F.~Augugliaro, A.~P. Schoellig, and R.~D'Andrea, ``Generation of collision-free
  trajectories for a quadrocopter fleet: A sequential convex programming
  approach,'' in \emph{2012 IEEE/RSJ International Conference on Intelligent
  Robots and Systems}.\hskip 1em plus 0.5em minus 0.4em\relax IEEE, 2012, pp.
  1917--1922.

\bibitem{luo2019importance}
Y.~Luo, H.~Bai, D.~Hsu, and W.~S. Lee, ``Importance sampling for online
  planning under uncertainty,'' \emph{The International Journal of Robotics
  Research}, vol.~38, no. 2-3, pp. 162--181, 2019.

\bibitem{morgan2016swarm}
D.~Morgan, G.~P. Subramanian, S.-J. Chung, and F.~Y. Hadaegh, ``Swarm
  assignment and trajectory optimization using variable-swarm, distributed
  auction assignment and sequential convex programming,'' \emph{The
  International Journal of Robotics Research}, vol.~35, no.~10, pp. 1261--1285,
  2016.

\bibitem{Luis(2019)}
C.~E. Luis and A.~P. Schoellig, ``Trajectory generation for multiagent
  point-to-point transitions via distributed model predictive control,''
  \emph{IEEE Robotics and Automation Letters}, vol.~4, no.~2, pp. 375--382,
  2019.

\bibitem{Park(2020)}
J.~Park, J.~Kim, I.~Jang, and H.~J. Kim, ``Efficient multi-agent trajectory
  planning with feasibility guarantee using relative bernstein polynomial,'' in
  \emph{2020 IEEE International Conference on Robotics and Automation
  (ICRA)}.\hskip 1em plus 0.5em minus 0.4em\relax IEEE, 2020, pp. 434--440.

\bibitem{Zhou(2020)}
X.~Zhou, J.~Zhu, H.~Zhou, C.~Xu, and F.~Gao, ``Ego-swarm: A fully autonomous
  and decentralized quadrotor swarm system in cluttered environments,'' in
  \emph{2021 IEEE International Conference on Robotics and Automation
  (ICRA)}.\hskip 1em plus 0.5em minus 0.4em\relax IEEE, 2021, pp. 4101--4107.

\bibitem{Ding(2019)}
W.~Ding, W.~Gao, K.~Wang, and S.~Shen, ``An efficient b-spline-based
  kinodynamic replanning framework for quadrotors,'' \emph{IEEE Transactions on
  Robotics}, vol.~35, no.~6, pp. 1287--1306, 2019.

\bibitem{chen2016pomdp}
M.~Chen, E.~Frazzoli, D.~Hsu, and W.~S. Lee, ``Pomdp-lite for robust robot
  planning under uncertainty,'' in \emph{2016 IEEE International Conference on
  Robotics and Automation (ICRA)}.\hskip 1em plus 0.5em minus 0.4em\relax IEEE,
  2016, pp. 5427--5433.

\bibitem{Tahir(2019)}
H.~Tahir, M.~N. Syed, and U.~Baroudi, ``Heuristic approach for real-time
  multi-agent trajectory planning under uncertainty,'' \emph{IEEE Access},
  vol.~8, pp. 3812--3826, 2019.

\bibitem{wolf2008artificial}
M.~T. Wolf and J.~W. Burdick, ``Artificial potential functions for highway
  driving with collision avoidance,'' in \emph{2008 IEEE International
  Conference on Robotics and Automation}.\hskip 1em plus 0.5em minus
  0.4em\relax IEEE, 2008, pp. 3731--3736.

\bibitem{Miao(2016)}
Z.~Miao, D.~Thakur, R.~S. Erwin, J.~Pierre, Y.~Wang, and R.~Fierro,
  ``Orthogonal vector field-based control for a multi-robot system
  circumnavigating a moving target in 3d,'' in \emph{2016 IEEE 55th Conference
  on Decision and Control (CDC)}.\hskip 1em plus 0.5em minus 0.4em\relax IEEE,
  2016, pp. 6004--6009.

\bibitem{Wang(2017)}
L.~Wang, A.~D. Ames, and M.~Egerstedt, ``Safety barrier certificates for
  collisions-free multirobot systems,'' \emph{IEEE Transactions on Robotics},
  vol.~33, no.~3, pp. 661--674, 2017.

\bibitem{Quan(2021)}
Q.~Quan, R.~Fu, M.~Li, D.~Wei, Y.~Gao, and K.-Y. Cai, ``Practical distributed
  control for {VTOL UAVs} to pass a virtual tube,'' \emph{IEEE Transactions on
  Intelligent Vehicles}, 2021.

\bibitem{khatib1986real}
O.~Khatib, ``Real-time obstacle avoidance for manipulators and mobile robots,''
  in \emph{Autonomous Robot Vehicles}.\hskip 1em plus 0.5em minus 0.4em\relax
  Springer, 1986, pp. 396--404.

\bibitem{panagou2014motion}
D.~Panagou, ``Motion planning and collision avoidance using navigation vector
  fields,'' in \emph{2014 IEEE International Conference on Robotics and
  Automation (ICRA)}.\hskip 1em plus 0.5em minus 0.4em\relax IEEE, 2014, pp.
  2513--2518.

\bibitem{panagou2016distributed}
------, ``A distributed feedback motion planning protocol for multiple unicycle
  agents of different classes,'' \emph{IEEE Transactions on Automatic Control},
  vol.~62, no.~3, pp. 1178--1193, 2016.

\bibitem{wang2016multi}
L.~Wang, A.~D. Ames, and M.~Egerstedt, ``Multi-objective compositions for
  collision-free connectivity maintenance in teams of mobile robots,'' in
  \emph{2016 IEEE 55th Conference on Decision and Control (CDC)}.\hskip 1em
  plus 0.5em minus 0.4em\relax IEEE, 2016, pp. 2659--2664.

\bibitem{hernandez2011convergence}
E.~G. Hern{\'a}ndez-Mart{\'\i}nez and E.~Aranda-Bricaire, \emph{Convergence and
  collision avoidance in formation control: A survey of the artificial
  potential functions approach}.\hskip 1em plus 0.5em minus 0.4em\relax INTECH
  Open Access Publisher Rijeka, Croatia, 2011.

\bibitem{rostami2019obstacle}
S.~M.~H. Rostami, A.~K. Sangaiah, J.~Wang, and X.~Liu, ``Obstacle avoidance of
  mobile robots using modified artificial potential field algorithm,''
  \emph{EURASIP Journal on Wireless Communications and Networking}, vol. 2019,
  no.~1, pp. 1--19, 2019.

\bibitem{antich2005extending}
J.~Antich and A.~Ortiz, ``Extending the potential fields approach to avoid
  trapping situations,'' in \emph{2005 IEEE/RSJ International Conference on
  Intelligent Robots and Systems}.\hskip 1em plus 0.5em minus 0.4em\relax IEEE,
  2005, pp. 1386--1391.

\bibitem{ge2005queues}
S.~S. Ge and C.~H. Fua, ``Queues and artificial potential trenches for
  multirobot formations,'' \emph{IEEE Transactions on Robotics}, vol.~21,
  no.~4, pp. 646--656, 2005.

\bibitem{vadakkepat2000evolutionary}
P.~Vadakkepat, K.~C. Tan, and W.~Ming~Liang, ``Evolutionary artificial
  potential fields and their application in real time robot path planning,'' in
  \emph{Proceedings of the 2000 Congress on Evolutionary Computation. CEC00
  (Cat. No. 00TH8512)}, vol.~1.\hskip 1em plus 0.5em minus 0.4em\relax IEEE,
  2000, pp. 256--263.

\bibitem{kim1992real}
J.~O. Kim and P.~Khosla, ``Real-time obstacle avoidance using harmonic
  potential functions,'' Ph.D. dissertation, Carnegie Mellon University, 1992.

\bibitem{rimon1990exact}
E.~Rimon, ``Exact robot navigation using artificial potential functions,''
  Ph.D. dissertation, Yale University, 1990.

\bibitem{panagou2015distributed}
D.~Panagou, D.~M. Stipanovi{\'c}, and P.~G. Voulgaris, ``Distributed
  coordination control for multi-robot networks using lyapunov-like barrier
  functions,'' \emph{IEEE Transactions on Automatic Control}, vol.~61, no.~3,
  pp. 617--632, 2015.

\bibitem{AIRBUS}
Airbus, ``Airbus skyways: the future of the parcel delivery in smart cities,''
  \url{https://www.embention.com/project/airbus-parcel-delivery/}, 2019.

\bibitem{Rasekhipour(2016)}
Y.~Rasekhipour, A.~Khajepour, S.~K. Chen, and B.~Litkouhi, ``A potential
  field-based model predictive path-planning controller for autonomous road
  vehicles,'' \emph{IEEE Transactions on Intelligent Transportation Systems},
  vol.~18, no.~5, pp. 1255--1267, 2016.

\bibitem{luo2018porca}
Y.~Luo, P.~Cai, A.~Bera, D.~Hsu, W.~S. Lee, and D.~Manocha, ``Porca: Modeling
  and planning for autonomous driving among many pedestrians,'' \emph{IEEE
  Robotics and Automation Letters}, vol.~3, no.~4, pp. 3418--3425, 2018.

\bibitem{Tony(2020)}
L.~A. Tony, A.~Ratnoo, and D.~Ghose, ``Corridrone: Corridors for drones, an
  adaptive on-demand multi-lane design and testbed,'' \emph{arXiv preprint
  arXiv:2012.01019}, 2020.

\bibitem{Liu(2017)}
S.~Liu, M.~Watterson, K.~Mohta, K.~Sun, S.~Bhattacharya, C.~J. Taylor, and
  V.~Kumar, ``Planning dynamically feasible trajectories for quadrotors using
  safe flight corridors in 3-d complex environments,'' \emph{IEEE Robotics and
  Automation Letters}, vol.~2, no.~3, pp. 1688--1695, 2017.

\bibitem{Likhachev(2004)}
M.~Likhachev, G.~J. Gordon, and S.~Thrun, ``Ara*: Anytime a* with provable
  bounds on sub-optimality,'' \emph{Advances in Neural Information Processing
  Systems}, vol.~16, pp. 767--774, 2003.

\bibitem{Dolgov(2010)}
D.~Dolgov, S.~Thrun, M.~Montemerlo, and J.~Diebel, ``Path planning for
  autonomous vehicles in unknown semi-structured environments,'' \emph{The
  International Journal of Robotics Research}, vol.~29, no.~5, pp. 485--501,
  2010.

\bibitem{Harabor(2011)}
D.~Harabor and A.~Grastien, ``Online graph pruning for pathfinding on grid
  maps,'' in \emph{Proceedings of the AAAI Conference on Artificial
  Intelligence}, vol.~25, 2011, pp. 1114--1119.

\bibitem{gammell2014informed}
J.~D. Gammell, S.~S. Srinivasa, and T.~D. Barfoot, ``Informed {RRT*}: Optimal
  sampling-based path planning focused via direct sampling of an admissible
  ellipsoidal heuristic,'' in \emph{2014 IEEE/RSJ International Conference on
  Intelligent Robots and Systems}.\hskip 1em plus 0.5em minus 0.4em\relax IEEE,
  2014, pp. 2997--3004.

\bibitem{gammell2015batch}
------, ``Batch informed trees {(BIT*)}: Sampling-based optimal planning via
  the heuristically guided search of implicit random geometric graphs,'' in
  \emph{2015 IEEE International Conference on Robotics and Automation
  (ICRA)}.\hskip 1em plus 0.5em minus 0.4em\relax IEEE, 2015, pp. 3067--3074.

\bibitem{Kavraki(1996)}
L.~E. Kavraki, P.~Svestka, J.~Latombe, and M.~H. Overmars, ``Probabilistic
  roadmaps for path planning in high-dimensional configuration spaces,''
  \emph{IEEE Transactions on Robotics and Automation}, vol.~12, no.~4, pp.
  566--580, 1996.

\bibitem{LaValle(1999)}
S.~M. LaValle and J.~J. Kuffner~Jr, ``Randomized kinodynamic planning,''
  \emph{The International Journal of Robotics Research}, vol.~20, no.~5, pp.
  378--400, 2001.

\bibitem{Gao(2019)}
F.~Gao, L.~Wang, K.~Wang, W.~Wu, B.~Zhou, L.~Han, and S.~Shen, ``Optimal
  trajectory generation for quadrotor teach-and-repeat,'' \emph{IEEE Robotics
  and Automation Letters}, vol.~4, no.~2, pp. 1493--1500, 2019.

\bibitem{quan2021far}
Q.~Quan, R.~Fu, and K.-Y. Cai, ``How far two {UAVs} should be subject to
  communication uncertainties,'' \emph{arXiv preprint arXiv:2110.09391}, 2021.

\bibitem{Slotine(1991)}
J.~J.~E. Slotine, W.~Li \emph{et~al.}, \emph{Applied nonlinear control}.\hskip
  1em plus 0.5em minus 0.4em\relax Prentice hall Englewood Cliffs, NJ, 1991,
  vol. 199.

\bibitem{Quan(2017)}
Q.~Quan, \emph{Introduction to multicopter design and control}.\hskip 1em plus
  0.5em minus 0.4em\relax Singapore: Springer, 2017.

\bibitem{Rezende(2020)}
A.~M. Rezende, V.~M. Gon{\c{c}}alves, A.~H. Nunes, and L.~C. Pimenta, ``Robust
  quadcopter control with artificial vector fields,'' in \emph{2020 IEEE
  International Conference on Robotics and Automation (ICRA)}.\hskip 1em plus
  0.5em minus 0.4em\relax IEEE, 2020, pp. 6381--6387.

\bibitem{rezende2018robust}
A.~M. Rezende, V.~M. Gon{\c{c}}alves, G.~V. Raffo, and L.~C. Pimenta, ``Robust
  fixed-wing {UAV} guidance with circulating artificial vector fields,'' in
  \emph{2018 IEEE/RSJ International Conference on Intelligent Robots and
  Systems (IROS)}.\hskip 1em plus 0.5em minus 0.4em\relax IEEE, 2018, pp.
  5892--5899.

\bibitem{Thomas(2009)}
G.~B. Thomas, R.~L. Finney, M.~D. Weir, and F.~R. Giordano, \emph{Thomas'
  calculus}.\hskip 1em plus 0.5em minus 0.4em\relax Addison-Wesley Reading,
  2003.

\bibitem{marsden2003vector}
J.~E. Marsden and A.~Tromba, \emph{Vector calculus}.\hskip 1em plus 0.5em minus
  0.4em\relax Macmillan, 2003.

\bibitem{tordesillas2019faster}
J.~Tordesillas, B.~T. Lopez, and J.~P. How, ``Faster: Fast and safe trajectory
  planner for flights in unknown environments,'' in \emph{2019 IEEE/RSJ
  International Conference on Intelligent Robots and Systems (IROS)}.\hskip 1em
  plus 0.5em minus 0.4em\relax IEEE, 2019, pp. 1934--1940.

\bibitem{borenstein1991vector}
J.~Borenstein, Y.~Koren \emph{et~al.}, ``The vector field histogram-fast
  obstacle avoidance for mobile robots,'' \emph{IEEE Transactions on Robotics
  and Automation}, vol.~7, no.~3, pp. 278--288, 1991.

\end{thebibliography}

\end{document}